\documentclass[10pt]{article}
\usepackage[utf8]{inputenc}
\usepackage{graphicx}
\usepackage{hyperref}

\usepackage{amsmath, amsthm, amsfonts, amssymb}

\baselineskip
\title{Towards a comprehensive visualization of structure in data}

\author{
Joan Garriga\textsuperscript{1}, Frederic Bartumeus\textsuperscript{1,2,3}
\\
\\
1 Theoretical and Computational Ecology Group,
\\ Centre d'Estudis Avan\c{c}ats de Blanes (CEAB-CSIC),
\\ \small{Cala Sant Francesc, 14, 17300, Blanes, Spain}
\\
2 Centre de Recerca Ecol\`ogica i Aplicacions Forestals (CREAF),
\\ \small{Cerdanyola del Vall\`es, 08193, Barcelona, Spain}
\\
3 Instituci\'o Catalana de Recerca i Estudis Avançats, ICREA, 
\\ \small{Passeig Lluís Companys, 23, 08010, Barcelona, Spain}
\\
E-mail: jgarriga@ceab.csic.es
\\
URL: \url{http://theelab.net/}
}

\date{}

\begin{document}

\maketitle

\begin{abstract}
Dimensional data reduction methods are fundamental to explore and visualize large data sets. Basic requirements for unsupervised data exploration are simplicity, flexibility and scalability. However, current methods show complex parameterizations and strong computational limitations when exploring large data structures across scales. Here, we focus on the t-SNE algorithm and show that a simplified parameter setup with a single control parameter, namely the perplexity, can effectively balance local and global data structure visualization. We also designed a chunk\&mix protocol to efficiently parallelize t-SNE and explore data structure across a much wide range of scales than currently available. Our parallel version of the BH-tSNE, namely pt-SNE, converges to good global embedding, comparable to state-of-the-art solutions, though the chunk\&mix protocol adds little noise and decreases the accuracy at the local scale. Nonetheless, we show that simple post-processing can efficiently restore local scale visualization, without any loss of precision at the global scales. We expect the same approach to apply to faster embedding algorithms other than BH-tSNE, like FIt-SNE or UMAP, thus, extending the state-of-the-art and leading to more comprehensive data structure visualization and analysis. 
\end{abstract}

\bigskip 
 
High dimensional data sets are prevalent in many areas of research. The computational analysis of these data often entails unsupervised exploratory steps where low dimensional visualization of data becomes crucial. Dimensional reduction techniques \cite{Gisbrecht:2015} are mainly divided into linear projections, focused on preserving the global structure of the data, \textit{e.g.} \textit{Principal Component Analysis} \cite{Hotelling:1993}, \textit{multidimensional scaling} \cite{Torgerson:1952} and non-linear embeddings, focused on preserving the local structure of the data, \textit{e.g.} \textit{Sammon mapping} \cite{Sammon:1969}, \textit{Isomap} \cite{Tenenbaum:2000}, \textit{Laplacian eigenmaps} \cite{Belkin:2001}. Two main ideas support the use of non-linear techniques for the visualization of high dimensional data: (i) high dimensional data is likely to be organized in a non-linear manifold of much lower dimension where the visualization of the global structure could make little sense, and (ii) linear projections of high dimensional data to a human-readable scale (i.e. two or three first components) might drop off essential local information. However, non-linear methodologies share serious drawbacks regarding the computational complexity of the algorithms and the interpretation of the output when the algorithm is not well-understood \cite{Wattenberg:2016}.

Two excellent methodologies stand out from the flurry of non-linear dimensional reduction techniques: (i) t-SNE (t-Stochastic Neighboring Embedding, \cite{Maaten:2008, Maaten:2009a}) with many faster variants (e.g. Barnes-Hut t-SNE (BHt-SNE) \cite{Maaten:2014}, MultiCore t-SNE (opt-SNE) \cite{Ulyanov:2016, Belkina:2019}, Fast-Fourier Interpolation-based t-SNE (FIt-SNE) \cite{Linderman:2019}) widely used in flow and mass cytometry and single-cell RNA sequencing data analysis (\cite{Wolf:2018, Belkina:2019}), and (ii) UMAP (Uniform Manifold Approximation and Projection, \cite{McInnes:2018}), increasingly prevalent in many areas of scientific research (\cite{Becht:2018}). t-SNE and UMAP are both gradient descent algorithms operating as a set of attractive and repulsive forces that make similar points attract each other and push dissimilar ones far away. Also, comparing UMAP and FIt-SNE, both algorithms are similar in terms of memory resources and computation times ($\mathcal{O}\left(n\right)$), both scaling well with large data sets.

Although t-SNE and UMAP are similar algorithms, there is an open discussion among practitioners about which algorithm is better at capturing the structure in the data \cite{Becht:2018, Kobak:2019}. In t-SNE and UMAP, a neighborhood-size parameter modulates how much the algorithm focus on capturing local or global structure, namely \textit{perplexity} (ppx) for t-SNE and \textit{nearest-neighbors} for UMAP. Based on this parameter, the first step in both algorithms is to compute a matrix of affinities, and the fact is that this step does not scale so well for large values of the neighborhood-size. Therefore, an important assumption in both frameworks is that the input data will usually reside in a manifold (a topological space where the neighborhood around every point is only locally Euclidean). While this assumption supports the use of low-ranged ppxs, it tells nothing about how large a locally Euclidean neighborhood can be. The effect of increased ppx is not yet thoroughly explored (just up to 0.001\% for a 1M data set \cite{Belkina:2019}), though the general assumption is that using low-ranged perplexities may constitute a potential limitation to unveiling higher levels of data structuring. Common techniques used to overcome this limitation are: (i) using informative initialization methods to inject the global structuring into the embedding from the very first stage of the algorithm (e.g. principal component analysis initialization, \cite{Maaten:2008}), and (ii) performing multi-scale analysis by mixing different scales (\cite{Lee:2015}, \cite{Kobak:2018}).

Another potentially limiting aspect of non-linear embedding techniques is the non-trivial parametric setup. The huge parameter space and the difficulty in understanding its influence on the output can be daunting to non-experts. In particular, for t-SNE, the main parameters are initialization, learning-rate, momentum, early/late exaggeration and ppx. In addition, a few more parameters control a contrived internal trickery used in the computation of the gradient descent to force a quick convergence and a neat visualization of the structure (e.g. momentum switch, early-exaggeration stop, step size update \cite{Jacobs:1988}). Such parametric complexity has motivated specific work giving some interesting clues though not compelling enough conclusions about optimal t-SNE parameterization\cite{Lee:2015, Wattenberg:2016, Cao:2017, Im:2018, Kobak:2018, Belkina:2019}. For the recently appeared UMAP, the lack of information about optimal tuning is even more critical.

We focus here on t-SNE with a double aim: first, to provide rationale on simplifying t-SNE parameterization, and second, to unleash the limits of t-SNE far beyond the range of small neighborhoods usually considered. For the latter objective, we assume that random subsets of data suffice in representing the whole data structuring. More generally, this would imply that large scale data sets convey large amounts of redundant evidence. Therefore, we propose to break down t-SNE into partial/parallel t-SNE runs on subsets of data, and we describe a chunk\&mix protocol that promotes the convergence of the partial solutions into a global one.

\section*{Results}

As a proof-of-concept, we provide a parallelized BHt-SNE, namely pt-SNE, where we apply the chunk\&mix protocol and a simplified parametric setup (see subsection~\nameref{sec:parametric_configuration} in Methods). Nevertheless, we want to highlight the generality of the concept, possibly suitable also to FIt-SNE or UMAP. We show that pt-SNE achieves good approximations to FIt-SNE, which we take here as our ground truth, and significantly reduces the computational complexity that penalizes t-SNE for increasing ppxs. Herein, we can explore data structuring at scales that otherwise remained prohibitive. The downside is a small cost in local accuracy that, if required, we can improve with simple post-processing later on.

We assess the goodness of pt-SNE visualizations by means of the \textit{k-ary neighborhood preservation} (kNP, \cite{Lee:2015}). The kNP depicts the matching between neighborhoods in the high and the low dimensional spaces (HD, LD) across a wide range of neighborhood sizes. A further summarization of kNP is the area under the curve (AUC, \cite{Lee:2015}). The kNP is commonly depicted against a logarithmic transformation of ppx to enhance the results of the low range of neighborhood sizes. As our interest extends now to higher values of ppx, we include a linear version of the kNP. Consequently, we distinguish linear and logarithmic versions of the AUC (linAUC and logAUC, respectively) to characterize the high-dimensional to the low-dimensional matching of global and local data structures.

\subsection*{Global/Local structure trade-off}

\begin{figure}[tp!]
\centering
\includegraphics[width=11.66cm, height=14.83cm]{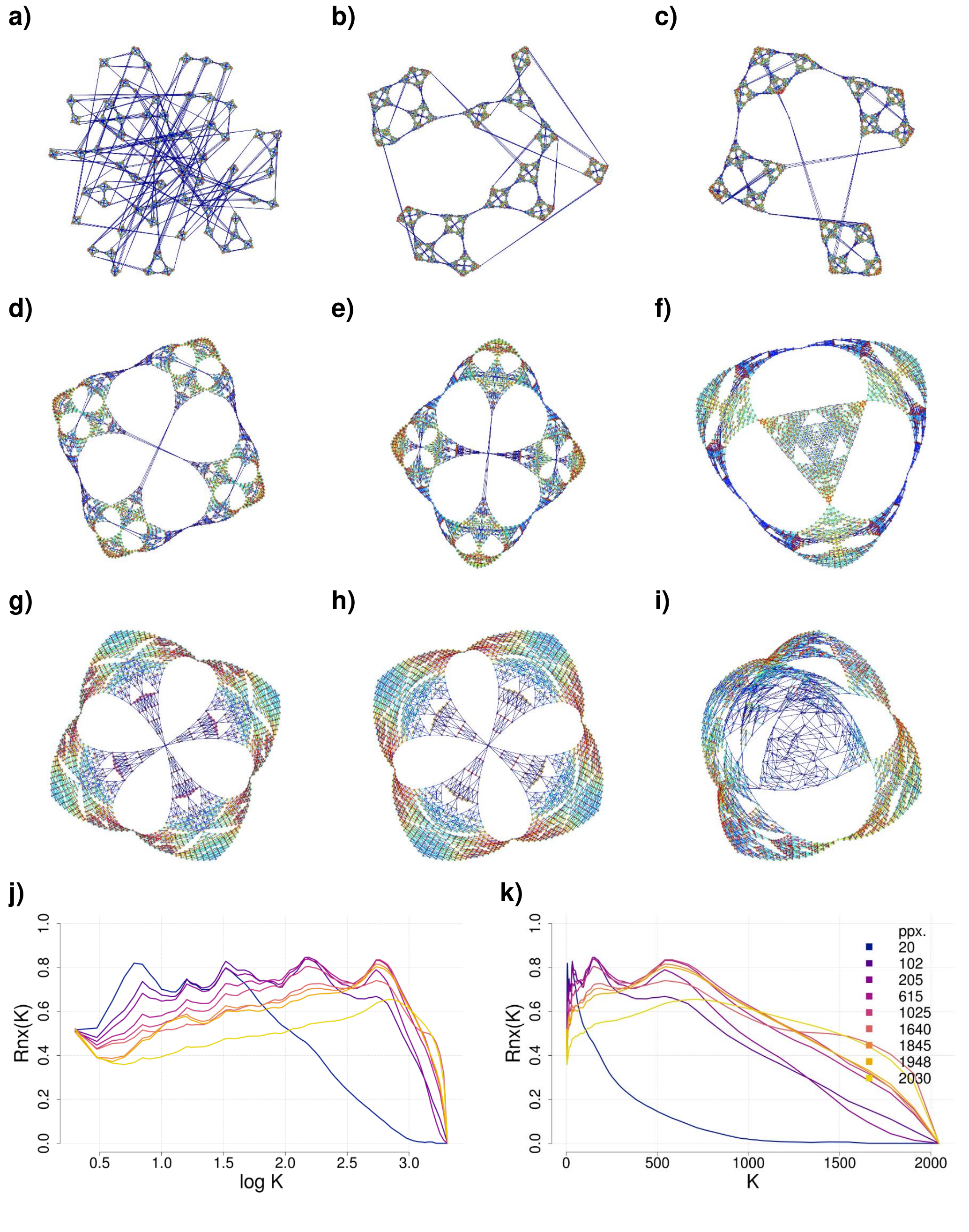}
\caption{\textbf{Global/Local trade-off} panels a) to i) pt-SNE visualization with ppx values corresponding to 0.01, 0.05, 0.10, 0.30, 0.50, 0.80, 0.90, 0.95 and 0.99 of the data set size. Colors depict relative pair-wise distances in original space (red:closer, blue:farther); j) log and k) linear kNP.}
\label{fig:s3d}
\end{figure}

It is well established that t-SNE fails to capture the global structure of large data sets \cite{Wattenberg:2016, Kobak:2018} and, since the appearance of UMAP, an increasing feeling among practitioners is that UMAP outperforms t-SNE both in output quality and running time \cite{Becht:2018}. This debate has been distorted so far by the inability to work with large neighborhood affinity matrices. Seemingly, the captured global structure is just the result of an informative initialization and can hardly be credited to any intrinsic advantage of either of the algorithms \cite{Kobak:2019}. We stress that t-SNE simultaneously optimizes both the local and the global structure in the data, and featuring more of the latter is just a matter of using large enough perplexities as the method prescribes. We show this by following the \textit{retrieval information} approach presented in \cite{Venna:2010}, where the authors introduce the \textit{neighbor retriever visualizer} (NeRV). NeRV is a visualization method that identifies the cost of retrieving/missing neighbors from the HD/LD representations of the data with the fundamental trade-off between \textit{precision} and \textit{recall} in information retrieval. By extending their approach (see Sec.~S1 in the Supplementary Material), we reach a reformulation of the t-SNE cost function that explicitly states how t-SNE assesses data structure at the local and global scale,

\begin{equation}
KL\left(P\| Q\right) \equiv \mathbb{E}_{p_{i}}\left[KL\left(p_{.\mid i}, q_{.\mid i}\right)\right] + KL\left(p_{i\mid .}, q_{i\mid .}\right)
\label{eq:qMeasure}
\end{equation}

The distributions $p_{.\mid i}$ and $q_{.\mid i}$ in Eq.~\ref{eq:qMeasure} are probabilistic models of \textit{smoothed recall}, i.e. describe the probability of picking a data point from the neighborhood of $\mathbf{x}_i$ in the HD and LD spaces, respectively. Analogously, the distributions $p_{i\mid .}\equiv p_{i}$ and $q_{i\mid .}\equiv q_{i}$ are probabilistic models of \textit{global prevalence}, i.e. represent the average probability of picking data point $\mathbf{x}_{i}$ as a neighbor of any other point. $KL\left(\right)$ is the Kullback-Leibler divergence. Thus, the first term in Eq. \ref{eq:qMeasure}, known as the \textit{expected smooth recall}, is an average measure of the mismatch between the HD and LD representations of the neighborhood of $\mathbf{x}_{i}$, weighted by the prevalence of $\mathbf{x}_{i}$. The second term is a measure of matching between the HD and LD models of prevalence, closely related to an index of global structure preservation. In summary, the t-SNE approach maximizes the \textit{expected smoothed recall} prioritizing areas where local structure is more significant while simultaneously trying to preserve the global structure. 

We illustrate this idea using a data set with a known structure (Fig.~\ref{fig:s3d}). The Sierpinski-3D data set is a graph representation of the well known Sierpinski 3D-triangle, where each row represents a node of the graph (a vertex of the structure) and each column is the \textit{shortest-path-distance} to the rest of the nodes (https://sparse.tamu.edu/, \cite{Hu:2005}, \cite{Kruiger:2017}). The Sierpinski-3D data set is not challenging in size but represents a challenging fractal structure for a dimensional reduction algorithm. By scanning this data set across a wide range of perplexities (Fig.~\ref{fig:s3d}) we observe how the algorithm accurately resolves the fractal structure at each specific scale (Fig.~\ref{fig:s3d} panels a to i, see also the Supplementary Material https://rpubs.com/bigMap/839757). Interestingly, the embedding starts showing the complete structure for ppxs beyond 30\% of the data (Fig.~\ref{fig:s3d}~f), an indication of the convenience of exploring high-ranged perplexities. Additionally, we include the kNP plots to depict a quantitative assessment of the embedding (Fig.~\ref{fig:s3d},~j, k). The kNP curves show how pt-SNE balances the prevalence of local and global structure signatures as ppx increases (Fig.~\ref{fig:s3d},~j) while the waving in each curve show how pt-SNE captures the inherent fractality of the object across different scales.

\subsection*{Speed/Accuracy trade-off}
\label{sec:speed_accuracy}

\begin{figure}[tp!]
\centering
\includegraphics[width=14.0cm, height=12.33cm]{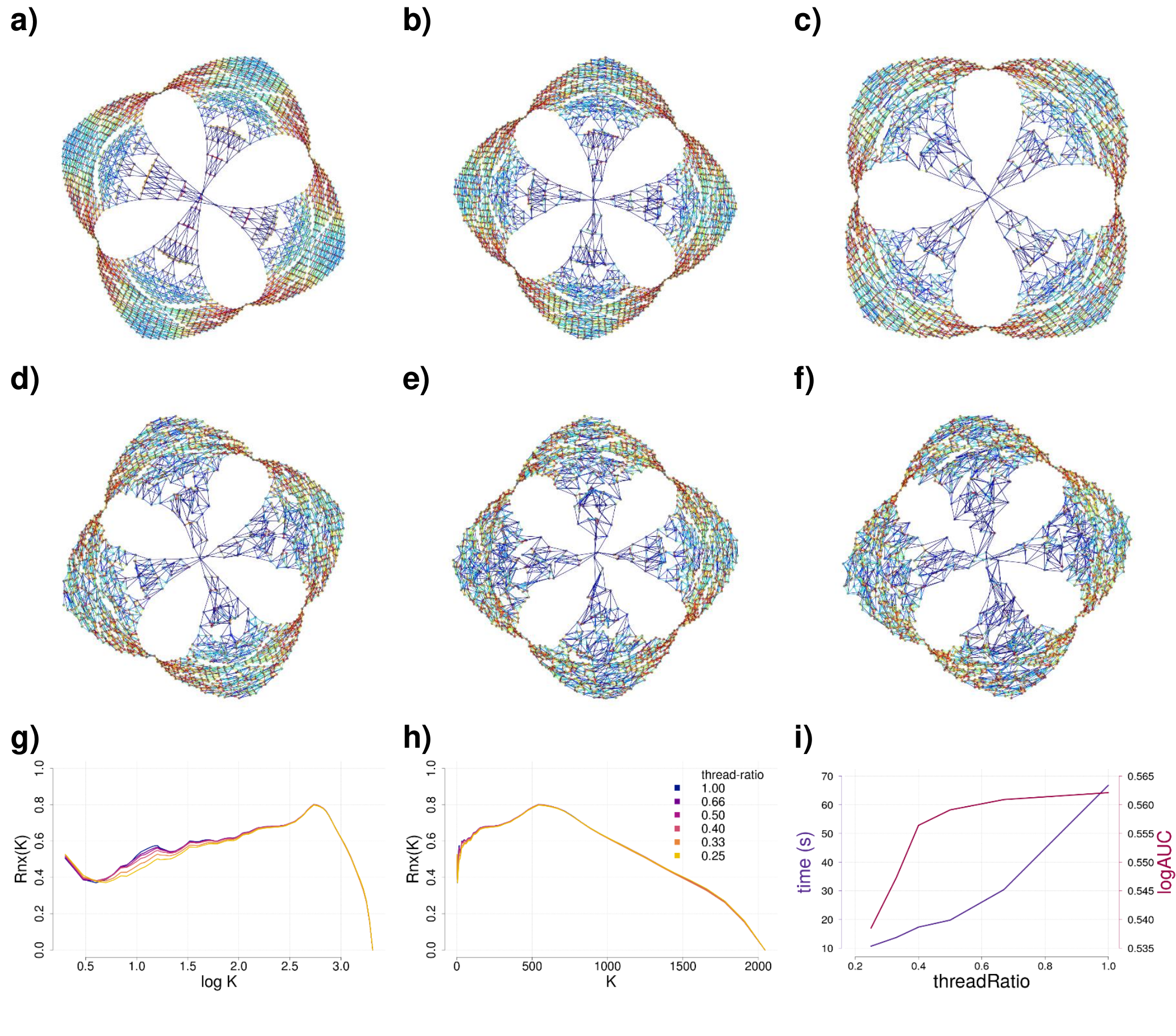}
\caption{\textbf{Speed/Accuracy trade-off} Panels a) to f) ptSNE visualization for $ppx = 1948$ and thread-ratio $\rho=\{1.0,\,0.67,\,0.5,\,0.40,\,0.33,\,0.25\}$; the local structure deteriorates as the thread-ratio decreases. Colors depict relative pair-wise distances in original space (red:closer, blue:farther); g) log and h) linear kNP; i) running-time and $logAUC$ versus thread-ratio.}
\label{fig:speedAcc}
\end{figure}

A fundamental parameter in the chunk\&mix protocol is the thread-ratio $\rho$ defining the proportion of data running in each partial t-SNE (see subsection~\nameref{sec:chunkandmix} in Methods). Using low thread-ratios, that is, splitting the data into more and smaller chunks, is the key to achieving reasonable running times at large perplexities. However, as the data chunks become smaller, the amount of structural information contained in each of them is lower, leading to a quality loss in the visualization of the local structure. This balance between chunk size and visualization quality is the speed/accuracy trade-off in pt-SNE. As an example, we show the visualization of the Sierpinski-3D data set for $ppx = 1948$ (same as in Fig.~\ref{fig:s3d}~h) with decreasing thread-ratio $\rho=\{1.0, 0.67, 0.50, 0.40, 0.33, 0.25\}$ (Fig.~\ref{fig:speedAcc} panels a to f, see also the Supplementary Material https://rpubs.com/bigMap/841359). As no subset of this data set can adequately reflect the overall data structure, a decrease in the \textit{thread-ratio} rapidly translates into a loss of information. In terms of kNP, a degradation of the local structure is clear (Fig.~\ref{fig:speedAcc}~g), but the global structure is preserved (Fig.~\ref{fig:speedAcc}~h). We also depict the speed/accuracy trade-off by plotting the running times and the logAUC versus the thread-ratio ranging from 1.0 down to 0.25 (Fig.~\ref{fig:speedAcc}~i).

\subsection*{Parametric setup}

Our parallel implementation of t-SNE demands a smooth parameter optimization avoiding too early cluster arrangements that may compromise the long run convergence among partial t-SNE solutions. To achieve such a smooth optimization pt-SNE shows a minimal parametric configuration where we drop \textit{exaggeration} and we use auto-adaptive schemes for \textit{momentum} and \textit{learning-rate} (see subsection.~\nameref{sec:parametric_configuration} in Methods). Furthermore, pt-SNE works well with random initialization. Therefore we are basically left with ppx (and the Barnes-Hut parameter $\theta$ which works well in general with the default value $\theta = 0.5$, \cite{Maaten:2014}). Additionally, the chunk\&mix protocol involves the \textit{thread-ratio} $\rho$, given by way of two parameters: $threads$, defining the number of chunks of data, and $layers$, defining the degree of overlapping among threads. These two parameters determine the proportion of data running in each partial t-SNE,  $\rho=layers/threads$,  (see subsection~\nameref{sec:chunkandmix} in Methods and Section~\nameref{sec:speed_accuracy}).

\begin{figure}[tp!]
\centering
\includegraphics[width=11.66cm, height=14.55cm]{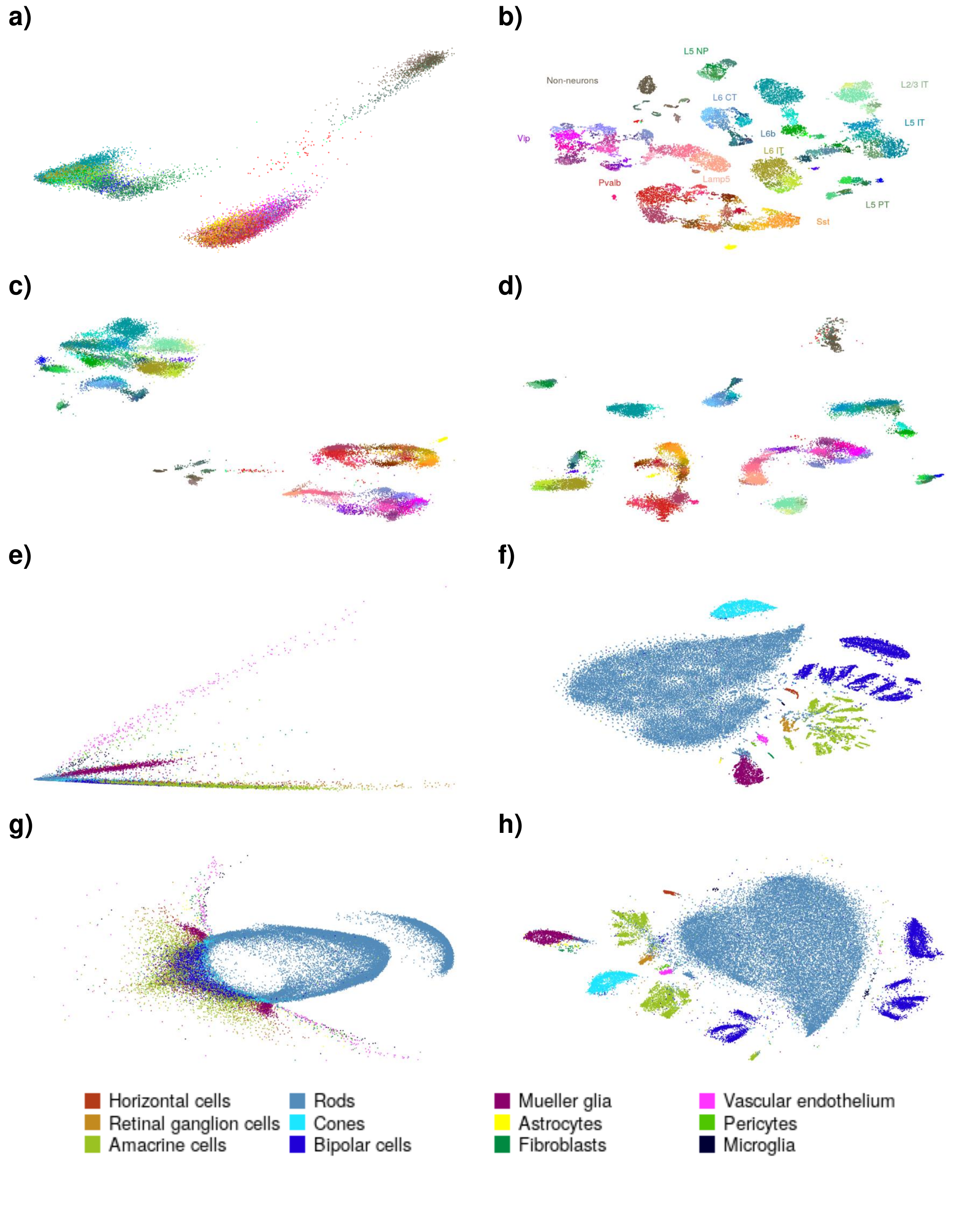}
\caption{\textbf{UMI-based transcriptomic data} Cluster assignments and annotations taken from the original publications. Panels a to d: \cite{Tasic:2018} $n = 23822$ cells from adult mouse cortex classified into 133 clusters.  Warm colors correspond to inhibitory neurons, cold colors correspond to excitatory neurons, brown/gray colors correspond to non-neural cells. Panels e to h: \cite{Macosko:2015} $n=44808$ cells from the mouse retina classified into 39 classes. Amacrine cells (green) and bipolar cells (dark blue) comprise 8 and 21 sub-classes respectively, not shown in the legend, but visible in the embedding). Panels a,~e: Global structure as depicted by the two first principal components of the data. Panels b,~f: FIt-SNE embedding as shown in \cite{Kobak:2018} (multi-scale affinities combining ppxs 30 and $n/100$). Panels c,~g: pt-SNE embedding using high perplexities (20\% and 40\% respectively). Panels d,~h: pt-SNE embedding using low perplexities (0.01\% and 0.005\% respectively).}
\label{fig:UMI}
\end{figure}

\cite{Kobak:2018} explains how to achieve improved t-SNE visualizations that preserve the global structuring in the data using real UMI-based transcriptomic data, e.g. \cite{Tasic:2018}, and \cite{Macosko:2015}. The main settings in \cite{Kobak:2018} included PCA initialization, multi-scale affinities combining perplexities 30 and $n/100$, learning-rate $eta=n/12$ and exaggeration $\alpha=12$. We run pt-SNE on the same data sets to check out our simple parametric setup involving random initialization, no exaggeration and auto-adaptive schemes for learning rate and momentum. Performing PCA on \cite{Tasic:2018} data (Fig. \ref{fig:UMI}~a) shows three well-separated groups corresponding to excitatory neurons (cold colors), inhibitory neurons (warm colors) and non-neural cells such as astrocytes or microglia (grey/brown colors). Performing separate PCA on these three subsets reveals further (local) structure in each of them (\cite{Kobak:2018}~Supplementary Material). \cite{Kobak:2018} showed that using their settings, FIt-SNE preserved much of the global structure in the data, clustering the different classes/sub-classes of cells in a coherent pattern. In the case of  \cite{Tasic:2018} data (Fig.\ref{fig:UMI}~b), inhibitory neurons are well-separated into two groups, Pvalb/SSt-expressing (red/yellow) and Vip/Lamp5-expressing (purple/salmon)). In the case of \cite{Macosko:2015} data (Fig:\ref{fig:UMI}~f), multiple clusters of amacrine cells (green), bipolar cells (dark-blue), and non-neural cells (classes from Mueller glia to microglia) are close together. Using pt-SNE, we achieve similar results by simply switching from high (Fig.~\ref{fig:UMI}~c,~g) to low (Fig.~\ref{fig:UMI}~d,~h) ppx values. Of note, pt-SNE does not mix multi-scale affinities but, instead, shows the different scales by gradually increasing the ppx (see the Supplementary Material https://rpubs.com/bigMap/840112 and https://rpubs.com/bigMap/840131). For instance, panels d and h show a similar level of local structure as reported in \cite{Kobak:2019} (panels b and f) but a lower level of the global one. However, pt-SNE can achieve higher levels of global structure (panels c and g), up to the point of improving the global structure suggested by the PC plots (panels a and e).

\subsection*{Computational scaling}

By breaking down the t-SNE into chunks, pt-SNE could potentially scale as much as the number of parallel t-SNE instances allowed by our hardware resources. On top of this, we designed pt-SNE to effortlessly work with high-performance computing (HPC) platforms using both intra and inter-node parallelization. However, two limitations arise: (i) decreasing the \textit{thread-ratio} in excess, that is, reducing too much the length of the data chunks, can result in a loss of data structure information and visualization over-blurring (Fig.~\ref{fig:speedAcc}, see also subsection~\nameref{sec:speed_accuracy}), and (ii) the chunk\&mix scheme (alternating short t-SNE runs with mixing of the partial solutions) adds an extra computational time (inter-epoch time) required to pool all partial t-SNE outputs after an epoch has finished, and this inter-epoch time increases with the number of partial t-SNEs (see subsection~\nameref{sec:chunkandmix} in Methods).

\begin{figure}[tp!]
\centering
\includegraphics[width=14.3cm, height=15.8cm]{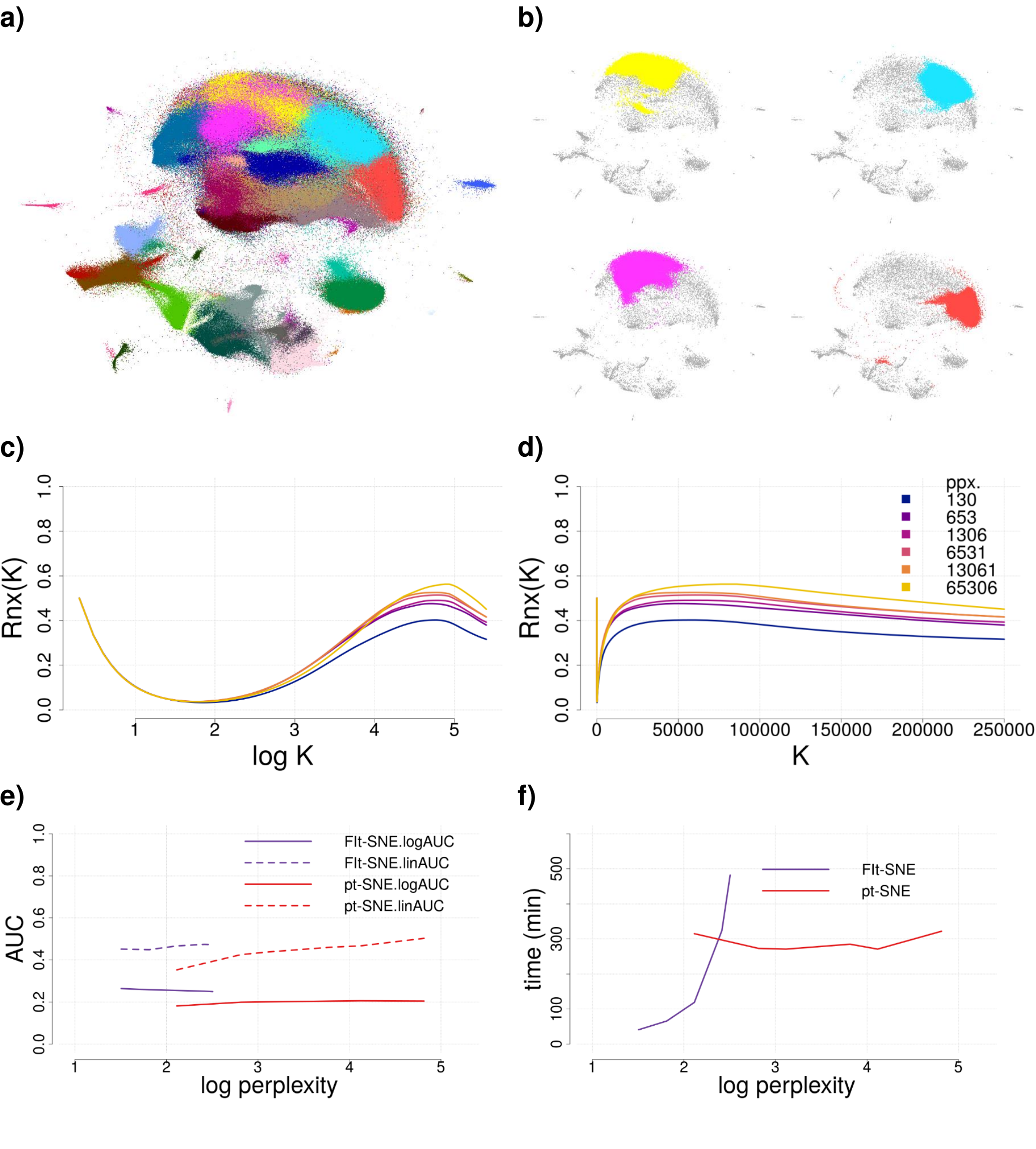}
\caption{\textbf{1.3 Million Brain Cells from E18 Mice} \cite{10xGenomics:2018}. Clustering labels taken from \cite{Wolf:2018}. a) pt-SNE embedding with $ppx=65306$ b) Overlay of the 4 largest classes; c),~d) log and linear kNP.; e),~f) pt-SNE vs. FIt-SNE, AUC and running times for different values of ppx.}
\label{fig:10xm}
\end{figure}

We run pt-SNE on the \cite{10xGenomics:2018} data set with $ppx=\{130, 653, 1301, 6531, 13061, 65306\}$, the latter equivalent to 5\% of the data set size (Fig.\ref{fig:10xm}) to show that, pt-SNE scales computationally for biological large data sets of about this size and up to perplexity values far beyond a few hundreds. The \cite{10xGenomics:2018} is a large transcriptomic data set ($n=1306127$ brain cells from E18 mice) commonly used to assess large scale visualization procedures (\cite{Wolf:2018, Kobak:2018, Belkina:2019}). The clustering labels for this data were derived by \cite{Wolf:2018} using the Louvain clustering algorithm (\cite{Blondel:2008}) and comprise 39 clusters. However, when visualized through dimensional reduction algorithms, these clusters show a significant overlap. Our analysis, with an augmented perplexity, offers an improved visualization in terms of compactness of the Louvain clusters (Supplementary Material https://rpubs.com/bigMap/841421) and in terms of kNP (both log and linear, Fig.\ref{fig:10xm}~c,~d), which could bring more accurate insights about the data. We run this analysis on an HPC platform using MPI and 130 cores (i.e. thread-ratio $\rho=0.015$, data chunks of 20094 observations). On this order of magnitudes, it took about 300 min (Fig.\ref{fig:10xm}~f) to complete the whole process (i.e. computing the bandwidths $\beta_i$ and running the gradient descent). We also run FIt-SNE for values of ppx ranging from 32 to 320 with quite similar results (Supplementary Material https://rpubs.com/bigMap/841412). We could not explore higher ranges because the amount of memory needed to run FIt-SNE with $ppx=320$ was already above 400GB. Of note, the ppxs in pt-SNE and FIt-SNE are not equivalent: in pt-SNE, the ppx must be higher to get similar results because the neighborhood size is smaller (see subsection~\nameref{sec:chunkandmix} in Methods). In terms of running times, FIt-SNE ($\mathcal{O}\left(n\right)$) is obviously faster than pt-SNE (indeed a BHt-SNE $\mathcal{O}\left(n\,\log\,n\right)$) but FIt-SNE is extremely dependent on ppx while pt-SNE is almost independent (Fig.\ref{fig:10xm}~f). FIt-SNE performed extremely well at capturing global structure from the very lowest value of ppx, only surmounted by pt-SNE at much larger values of ppx (Fig.\ref{fig:10xm}~e). However, pt-SNE embedding shows higher sensitivity to ppx than FIt-SNE (note the increasing linAUC for increasing ppxs, dashed red line in panel e). Therefore, pt-SNE highlights much better the differences of data organization across scales. The drawback is a loss of local structure (solid red line in panel e) due to the small size of the data chunks, (i.e. low thread-ratio $\rho=0.015$).  

We also compared pt-SNE and FIt-SNE on the \textit{Primes-1M} data set~\cite{Williamson:2019}. This data set is a structured representation of the first $10^6$ integers based on their prime factor decomposition, conforming to a highly sparse matrix with 78498 dimensions (the set of primes lower than $10^6$) where the values are the prime factorization powers. We note some differences: (i) in Williamson’s work, any factor is represented as either a zero or a one, just indicating divisibility by that factor, while we use the actual power for each factor to have unique representations for each integer; (ii) Williamson computes cosine-similarity based affinities, while we use euclidean-distance based affinities.

We run pt-SNE on the \textit{Primes-1M} dataset with $ppx=\{1000, 5000, 10000, 20000, 50000\}$ (equivalent to a neighborhood size of 0.1, 0.5, 1, 2 and 5\% of the data set size, (see Supplementary Material in https://rpubs.com/bigMap/840981). We run pt-SNE using MPI and 100 cores with 4GB/core (thread-ratio $\rho=0.02$). We also run FIt-SNE on \textit{Primes-1M} data set with $ppx=\{50, 100, 200, 300, 400\}$ (Supplementary Material https://rpubs.com/bigMap/840980). As Fit-SNE in~\cite{klugerlab:2019} does not have support for sparse matrices, for this dataset we used FIt-SNE from OpentSNE \cite{Policar:2019}, using a single node with 20 cores and 400GB.

\begin{figure}[tp!]
\centering
\includegraphics[width=13.00cm, height=11.45cm]{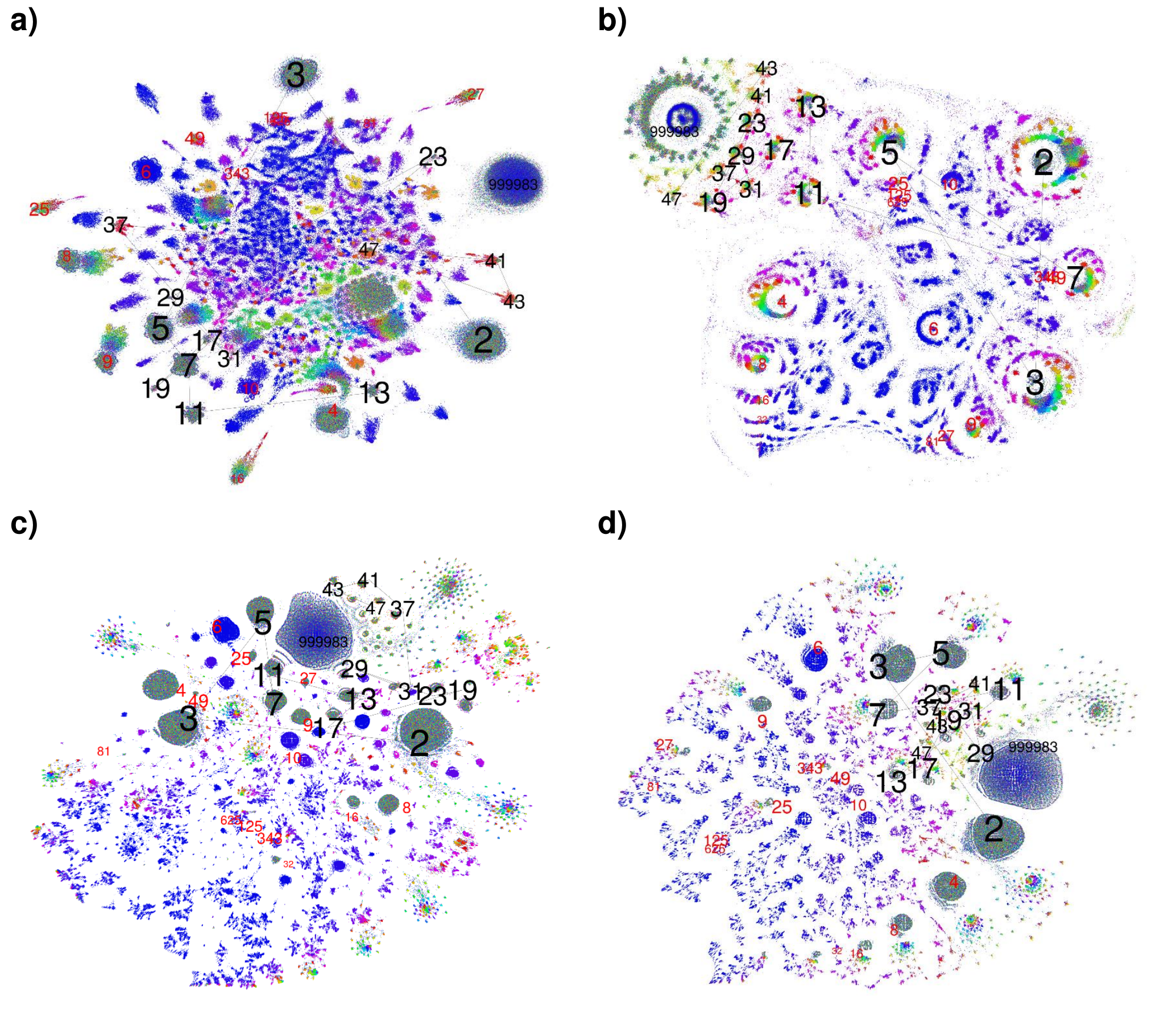}
\caption{\textbf{Primes-1M data set embedding}. a), b) pt-SNE visualization with $ppx=\{1000, 50000\}$; c), d) FIt-SNE visualization with $ppx = \{50, 400\}$. Hue color component is a combination of the two first prime factors while saturation depict the factors' powers. Embedding position of the first 15 primes (2 to 47) and of the last prime (999983) displayed in black. Embedding positions of some primes' powers of interest (e.g. 4, 8, 16, 32) displayed in red.}
\label{fig:P1G_embedding}
\end{figure}

\begin{figure}[tp!]
\centering
\includegraphics[width=13.00cm, height=12.90cm]{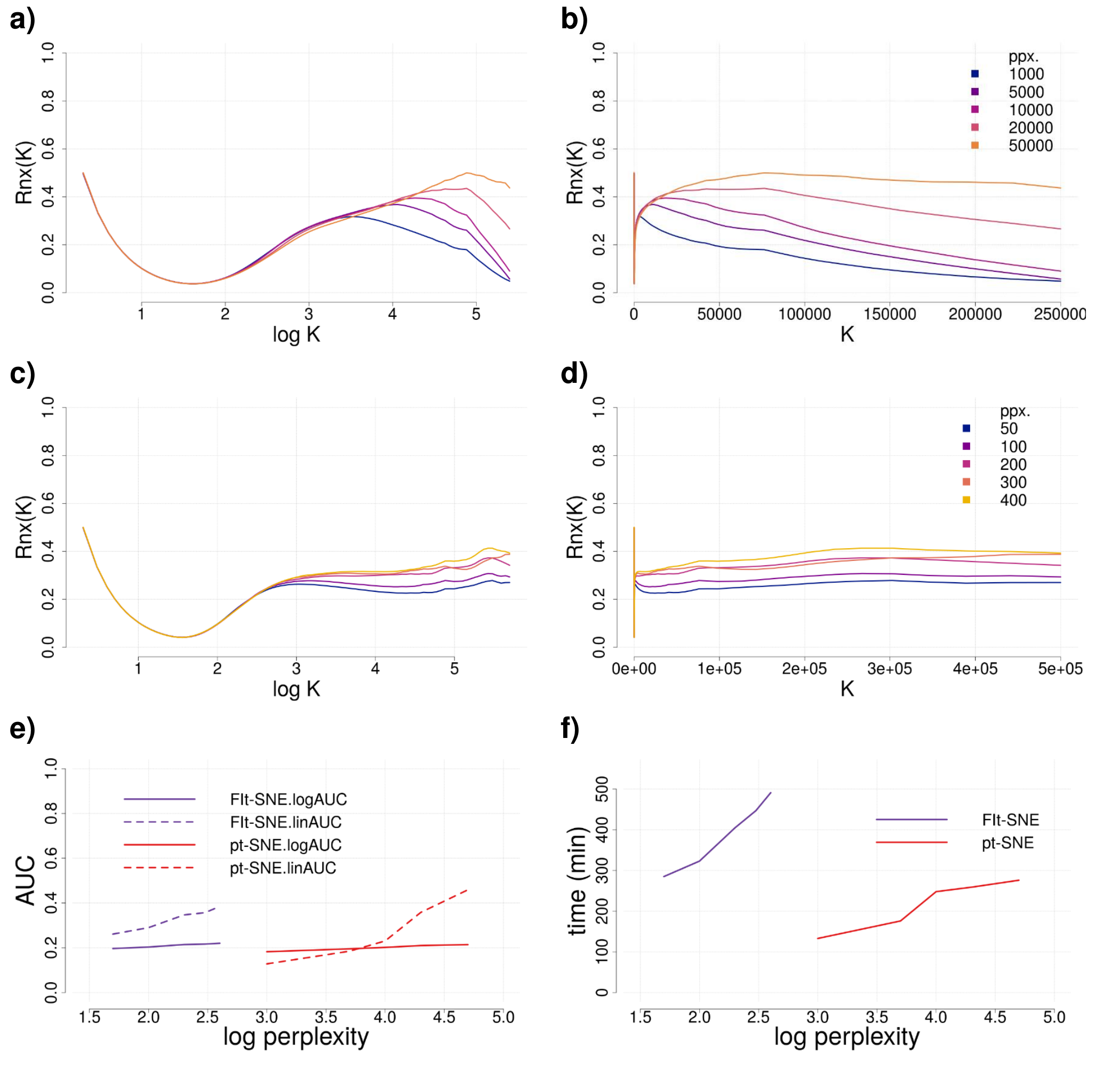}
\caption{\textbf{Primes-1M data set. ptSNE embedding versus FIt-SNE embedding}. a), b) pt-SNE log and linear kNP; c), d) FIt-SNE log and linear kNP; e) pt-SNE vs. FIt-SNE logAUC; f) pt-SNE vs. FIt-SNE running times.}
\label{fig:P1G_plots}
\end{figure}

The~\textit{Primes-1M} showed a complex data structure with strong hierarchical relations between its components not equally depicted by both algorithms (Fig. \ref{fig:P1G_embedding}). The differences observed between the pt-SNE and the Fit-SNE visualizations are mainly due to a different computation of the bandwidths $B_{i}$ (see Supplementary Material Section~S2.3). Also, the two algorithms showed unequal behaviour as we moved from local to global scales. On the one hand, FIt-SNE returned a quite neat depiction of the structure, capturing much of both the local and the global structure, even from the lowest value of ppx (\ref{fig:P1G_embedding}~c). Increased ppxs slightly improved the global structure without any loss of information at the local level (Fig. \ref{fig:P1G_plots}~c,~d). However, the embedding landscape did not change significantly overall (\ref{fig:P1G_embedding}~c,~d). On the other hand, beyond $ppx=10000$ pt-SNE starts revealing similar structures as those shown by FI-tSNE, but adding the characteristic blurring due to the chunk\&mix protocol (Fig. \ref{fig:P1G_embedding}~a,~b). Noteworthy, pt-SNE embedding outputs were much more sensitive to ppx than the ones derived from FIt-SNE (compare Fig. \ref{fig:P1G_plots}~a,~b) with Fig. \ref{fig:P1G_plots}~c,~d), revealing novel details at high perplexities not observed with FI-tSNE. We can also assess the higher sensitivity of pt-SNE landscapes to ppx by looking at the linAUC curves (global structure) in Fig. \ref{fig:P1G_plots}~e (doted-lines). By means of pt-SNE, we \textit{gradually} evolve the landscape from local (Fig. \ref{fig:P1G_embedding}~a) to global structure (Fig. \ref{fig:P1G_embedding}~b) by simply increasing the ppx (see Supplementary Material https://rpubs.com/bigMap/840981), and this is just the aim of pt-SNE. In terms of running times, pt-SNE shows a lower dependence on ppx and lower running times than FIt-SNE as implemented in OpentSNE (Fig. \ref{fig:P1G_plots}~f).

Both examples in this section, the first from the biological domain, the second from the mathematical one, illustrate the main features of pt-SNE: (i) independence of running times to ppx, enabling global data structure analysis beyond the state-of-the-art (ii) high sensitivity of embedded landscapes to ppx, gradually depicting data organization across structuring scales, and (iii) blurring at local scales due to the chunk\&mix protocol, highlighting an unavoidable trade-off between computational speed and accuracy. In section~S3 of the Supplementary Material, we explain how simple post-processing can efficiently restore the accuracy of the local structure while keeping the global data structure obtained with pt-SNE (Figures S1 and S2).

\section*{Discussion}

As a first principle, methods for data exploratory analysis should be as simple and flexible as possible. Nonetheless, it is not easy to accommodate and fulfill these principles, hence, key conceptual and operational limits exist in current visualization methods. FIt-SNE \cite{Linderman:2019} and UMAP \cite{McInnes:2018} are both outstanding algorithms that can perform the gradient-descent on large data sets within very reasonable times but show complex parameterizations and computational limitations to explore large scale data structures. Focused on t-SNE, our work aims at breaking through these two limitations.

At the origin of the parametric complexity of t-SNE there is a contrived trickery in the computation of the gradient descent intended to force a quick convergence and a neat depiction of the structure. We noted that the gradient descent equation suggests by itself an auto-adaptive scheme for the learning-rate (Eq.~\ref{eq:learning_rate}) that smoothly drives the embedding to an optimal solution. Based on this observation, we showed that a simple parametric setup, using random initialization, auto-adaptive schemes for learning-rate and momentum and discarding the use of exaggeration can achieve similar results as those reported in the related literature.

The limitation at exploring data across scales might stem from the t-SNE heuristic itself or the algorithmic implementation of the heuristic. We showed that the t-SNE heuristic simultaneously optimizes both local and global structure (Eq.~\ref{eq:qMeasure}) and that the final embedding is a balanced visualization of one or the other only depending on the value of ppx, just as the heuristic prescribes. Thus, the actual limitation to explore data across scales is purely an implementation issue stemming from both data set size and the neighboring size parameter (i.e. ppx). To overcome this limitation, we introduced a chunk\&mix protocol as a parallelized re-implementation of t-SNE. The underlying assumption is that large data sets convey a lot of redundant evidence so that random subsets of data comprehensively reflect the structure in the data. Therefore, we can approximate the solution by running instances of the algorithm on random subsets of the data and combining the partial results. This approach extends the state-of-the-art, overcoming the impossibility of computing the affinity matrix for large values of ppx, and allows scanning data structure across a wide range of scales. The downside is a loss of detail in the definition of the local structure, the loss increasing as the data chunks are shorter or the redundancy in the data is less.

As a proof of concept, we developed pt-SNE, a parallel version of BHt-SNE, and we showed that the chunk\&mix protocol converges to good global embedding. Hence, we expect the same approach to apply to other visualization algorithms (e.g. FIt-SNE or UMAP). pt-SNE runs efficiently on HPC platforms up to $10^6$ observations. To scale beyond this order of magnitude, we can either increase the number or the size of the data chunks. Increasing the number of data chunks may result in too small data subsets and excessive loss of information at local scales. Still, we could overcome this accuracy loss by combining a large ppx pt-SNE with a subsequent dimensional reduction algorithm (e.g. FIt-SNE) with lower ppx, using the former to sketch the global structure and the latter to refine the local one. Increasing the size of the data chunks results in a quadratic increase of running time (being pt-SNE based on BHt-SNE), so it is not a practical solution. However, algorithms like FIt-SNE and UMAP around $\mathcal{O}\left(n\right)$ could help improve this limitation. Therefore, a promising solution is to apply the chunk\&mix approach with these much faster algorithms, thus boosting the computation of the gradient descent and allowing larger thread sizes within reasonable running times.

\section*{Methods}

The pt-SNE runs multiple instances (independent threads) of the t-SNE on different chunks of data (partial t-SNEs). The algorithm starts by randomly allocating the data points in the low-dimensional space (a 2D half-unit disk, i.e. of radius $r=0.5$). Afterwards, a cyclic scheme arranged into \textit{epochs} optimizes the embedding (Fig. \ref{fig:ptSNE_scheme1}), each epoch involving: shuffling and chunking the data set indexes, exporting a chunk of indexes to each thread, running the partial t-SNEs with a short number of iterations, and pooling the solutions of the partial t-SNEs into a global embedding. 

\subsection*{Chunk\&Mix parameters}
\label{sec:chunkandmix}

The parameters that define the chunk\&mix scheme are the following:

\begin{figure}[!t]\centering
\includegraphics[width=13.5cm, height=6.5cm]{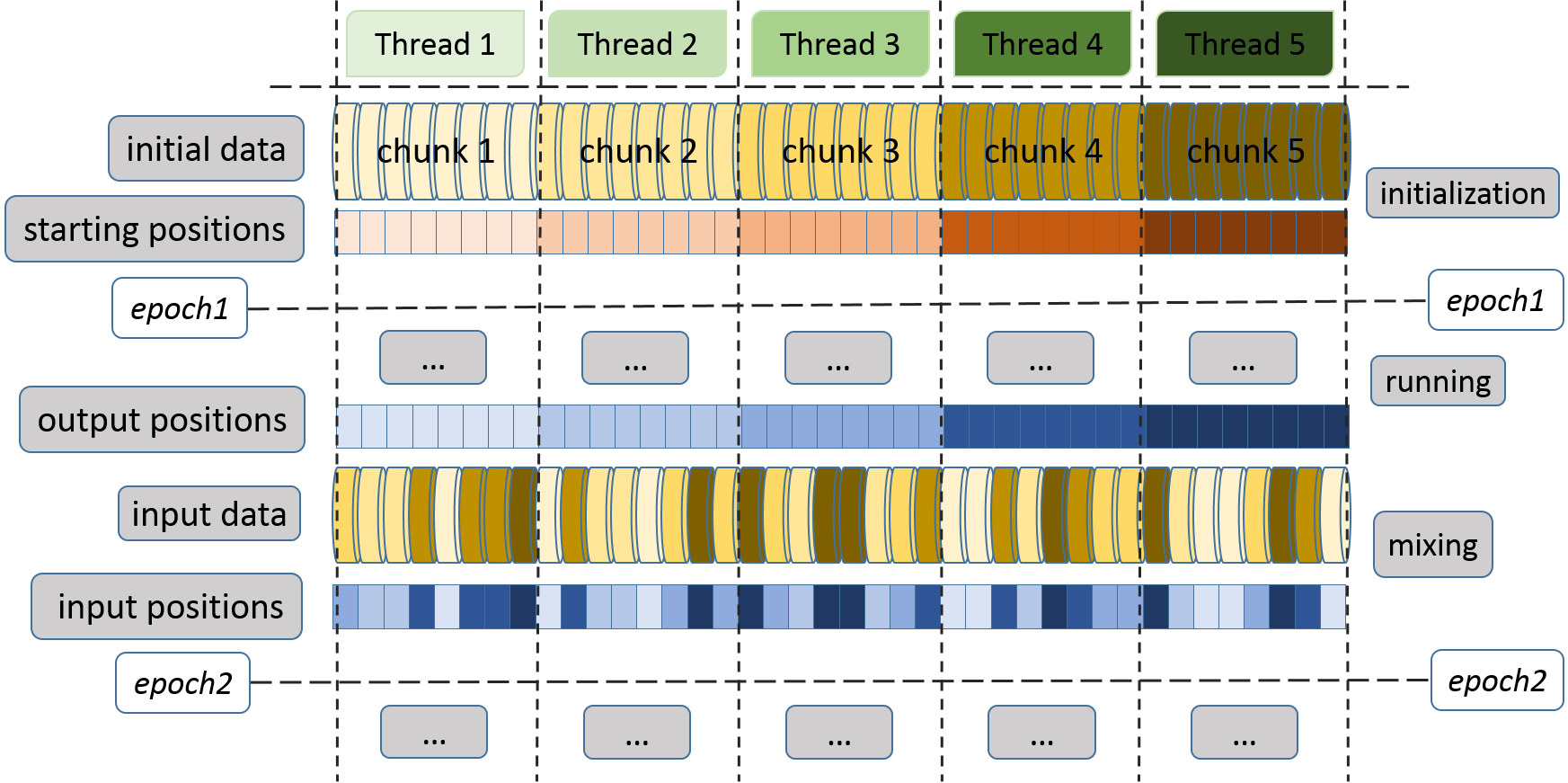}
\caption{\textbf{ptSNE basic parallelization scheme}. The parallelized implementation runs a number (5 in this example) of instances of the t-SNE algorithm in an alternating scheme of short runs and mixing of partial solutions. Each run-and-mix phase is an epoch. In this example, each thread iterates on a single chunk of data, starting with random mapping positions. After a number of iterations the partial t-SNEs are pooled together and mixed. A new epoch is started with each thread iterating on a new chunk of data and its current mapping positions.}
\label{fig:ptSNE_scheme1}
\end{figure}

\begin{figure}[!th]\centering
\includegraphics[width=12cm, height=5.5cm]{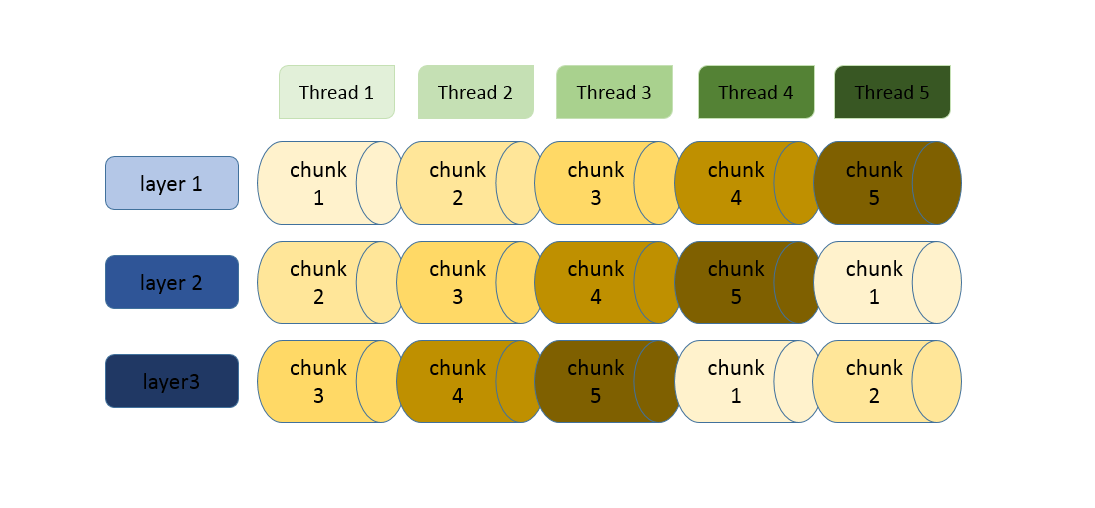}
\caption{\textbf{ptSNE parallelization scheme with 3 layers}. Each thread iterates on 3 chunks of data sharing each one of them with successive threads. Common data points create a link between the partial solutions that favors convergence. As each data point is running on 3 different threads we get 3 different mapped positions for each one. After pooling all partial solutions we get 3 global mapping layers.}
\label{fig:ptSNE_scheme3}
\end{figure}

\paragraph{threads}
The number of threads $z$ is the number of partial t-SNEs that will run. The pt-SNE splits the data set into this number of elementary chunks, so that, the larger the number of threads, the faster the computation of the final solution. Note that the number of threads can be higher than the number of physical cores available (what is known as multi-threading). Using multi-threading can yield further reductions in computational time.

\paragraph{layers}
In the most simple scheme (Fig. \ref{fig:ptSNE_scheme1}), each thread runs a single chunk of data. However, the key to convergence to a common solution is to impose some overlap among data chunks. Instead of running single chunks of data, threads are cyclically chained so that each thread includes at least two chunks of data. Thus, each thread shares one-half of the data points with the previous thread and the other half with the next one. The number of \textit{layers} sets the degree of overlapping. We name this parameter \textit{layers} because the overlapping makes each data-point take part in at least two partial t-SNEs and, after pooling all partial solutions, we will have at least two layers of global solutions. As an example, setting $threads=5$ and $layers=3$ (Fig. \ref{fig:ptSNE_scheme3}) the pt-SNE pools chunks 1, 2, and 3 into thread 1, chunks 2, 3 and 4 into thread 2, and so on, up to chunks 5, 1 and 2 into thread 5).

\paragraph{thread-ratio and thread-size}
The \textit{thread-ratio} is defined as $\rho=layers/threads$ and determines the thread-size $\nu=\rho\,n$. Being t-SNE of order quadratic to the size of the data set, making $\nu\ll n$ overcomes the unsuitability of the t-SNE algorithm for large-scale data sets. The thread-ratio represents a trade-off between accuracy and computational time. The closer is the thread-ratio to 1, the more robust and comprehensive is the solution, but the larger the computational cost. However, for large data sets ($n>10^4$) where the redundancy assumption holds, pt-SNE yields a good global solution even with values of $\rho$ as low as 0.01.

\paragraph{epochs and iterations}
The number of epochs is set to $2\,log\,n^2$ (where $n$ is the data set size) and the number of iterations per epoch is set to $2\,log\,\nu^2$ (where $\nu$ is the \textit{thread-size}). We empirically determined that these settings allow reaching a stable solution. Nevertheless, the user can restart the process and run it for additional epochs if the cost function does not show a flat line (Fig.~\ref{fig:gradient_descent}~b,~c). The epoch running time depends on the number of iterations performed by the threads, plus an inter-epoch time for the master process to pool the solutions, mixing them and sending new chunks to the workers.

\subsection*{Partial t-SNEs}

The t-SNE algorithm starts by transforming similarities (usually given as pairwise euclidean distances) into a probability distribution known as the \textit{affinity matrix}. Computing the whole affinity matrix is prohibitive for large data sets. The pt-SNE splits the data set into chunks and runs parallel instances of the t-SNE with subsets of the data, each partial t-SNE computing a much smaller affinity matrix. When splitting a data set $X$ into $z$ chunks, we denote the subset of data in each partial t-SNE as $X^k,\, 1\leq k\leq z$, and we denote the neighborhood of $i$ in $X$ as $N_{i}$ and the neighborhood of $i$ in $X^{k}$ as $N^{k}_{i}$. Thus, in each partial t-SNE, we compute the similarities in the input and output spaces as follows:

\paragraph{Similarities in the input (high dimensional) space, $\mathcal{X}\in \mathcal{R}^m$}

The similarity between observations $x_j$ and $x_i$, expressed as $\|x_i-x_j \| ^2$, is converted into the conditional probability $p_{j\mid i}$ given by a Gaussian kernel centered at $x_{i}$,

\begin{flalign*}
p_{j\mid i} = \frac{\exp\left(-\beta_{i}\,\|x_i-x_j\| ^2\right)}{\sum_{k\neq i}\exp\left(-\beta_{i}\,\|x_i-x_k\|^2\right)}
&\;,\quad i\in X^{k},\; j\in N^{k}_{i}
\\
p_{j\mid i} = 0
&\;,\quad i\in X^{k},\; j\notin N^{k}_{i}
\end{flalign*}

\noindent with precision $\beta_{i}=1/\left(2\,\sigma_i^2\right)$. Decreasing values of $\beta_{i}$ induce a probability distribution of increasing entropy $H\left(p_{j|i}\right)$ and increasing \textit{perplexity}, defined as,

\begin{equation*}
ppx\left(p_{j|i}\right)=2^{H\left(p_{j|i}\right)}\;,\quad i\in X,\; j\in N_{i}
\label{eq:perplexity}
\end{equation*}

\noindent We stress that $\beta_{i}$ are computed globally, i.e. for the whole data set. The maximum perplexity is equal to $n-1$ (where $n$ is the data set size), corresponding to $\beta_{i}=0$ and a uniform affinity distribution. In a preliminary step, t-SNE computes the values $\beta_{i}$ that result in a fixed perplexity for all $x_{i}$. We describe the procedure to find $\beta_{i}$ in Supplementary File S1. Computing similarities based on perplexities is a powerful transformation because it allows defining affinities in terms of spatial proximity without explicitly referring to any actual distance. The usual interpretation of perplexity is the number of neighbors to be picked: low perplexity will unveil the local structure in the data, whereas high perplexity will enhance the emergence of global structuring. Thus, it is fundamental to tune perplexity according to our requirements or, alternatively, explore data structuring across a range of perplexities. Afterwards, each partial t-SNE computes a symmetric joint probability given by,

\begin{equation*}
p_{ij} = p_{ji} = \frac{p_{j\mid i}+p_{i\mid j}}{2\nu}
\label{eq:h_joint_prob}
\end{equation*}

\noindent where $\nu$ is the thread-size, thus $\sum_{i,j}\,p_{ij} = 1$, and $\sum_j\,p_{ij}>\frac{1}{2\nu},\;\forall\,x_{i}$, so that each data point plays its role in the embedding process \cite{Maaten:2008}.

\paragraph{} To reduce the complexity in the computation of attractive forces, \cite{Maaten:2014} proposed to use a neighborhood size of $N_i=3\,ppx$, dramatically decreasing the size of the affinity matrix. To find the nearest neighbors' sets, standard implementations of t-SNE make use of fast approximations like \textit{vantage point trees} \cite{Nielsen:2009} or ANNOY \cite{Annoy:2018} and, afterwards, determine the local bandwidths and compute the affinity matrix. In pt-SNE, the partial t-SNEs run a different chunk of data at each new epoch, thus requiring the recomputation of the affinity matrices. We alleviate this process by precomputing the global neighborhoods $N_i$ and the bandwidths $\beta_i\left(N_i\right)$, and we pull out the distance of the furthest neighbor $L_i=max\{Lij\mid\, j\in N_i\}$. The values $B_i$ and $L_i$ are shared to all processes so that each one can use it to determine the neighborhoods $N^k_i=\{j\in\,X^k\mid\,L_{ij}\ll L_i\}$ and compute the partial affinity matrix. This process adds a short inter-epoch computation time. On average, $N^k_i\approx\rho\,3\,ppx$, thus, for low values of $\rho$, the number of forces acting on a data point is much lower in pt-SNE than in standard t-SNE. Therefore, as a general rule, we will use higher perplexities in pt-SNE to have equivalent outputs.

\paragraph{Similarities in the output (low-dimensional) space, $\mathcal{Y}\in \mathcal{R}^d$, $d\in\{2, 3\}$}

The similarities between mapped data points $y_j$ and $y_i$, also expressed as $\|y_i - y_j\|^2$, are treated differently. A well-known issue of embedding processes is the so-called \textit{crowding problem} (i.e. a surface at a given distance from a data point in a high-dimensional space can enclose more data points than those that fit in the corresponding low-dimensional area \cite{Maaten:2008}). This problem is alleviated using a heavy-tailed distribution to represent affinities in the low dimensional space, namely a Cauchy distribution (i.e. a t-Student distribution with one degree of freedom). Therefore, we define the joint probabilities $q_{ij}$ as,

\begin{equation}
q_{ij} = \frac{\left(1 + \| y_i-y_j \|^2\right)^{-1}}{\sum_{k\neq l}\left(1 + \| y_k-y_l \|^2\right)^{-1}}\;,\quad i,\,j \in X^{k}
\label{eq:l_joint_prob}
\end{equation}

\subsection*{Cost function}

t-SNE uses a gradient descent method to find a low-dimensional representation of the data that minimizes the mismatch between $p_{ij}$ and $q_{ij}$. The cost function is defined as the Kullback-Leibler divergence between both distributions (\cite{Maaten:2008}),

\begin{equation}
\label{eq:cost_function}
C =KL\left(P\|Q\right)=\sum_{i,j}p_{ij}\log\frac{p_{ij}}{q_{ij}}
\end{equation}

\noindent with a gradient with respect to the low-dimensional mapped positions given as (\cite{Maaten:2008}),

\begin{equation}
\frac{\delta C}{\delta y_i} = 4\sum_j\left(p_{ij}-q_{ij}\right)\left(y_i-y_j\right)\left(1+\|y_i-y_j\|^2\right)^{-1}
\label{eq:cost_gradient}
\end{equation}

\noindent The mode the gradient descent operates is clear by noting $L_{ij} \equiv \frac{\left(y_i-y_j\right)}{\left(1+\|y_i-y_j\|^2\right)}$, and writing Eq.\ref{eq:cost_gradient} as,

\begin{equation}
\frac{1}{4}\frac{\delta C}{\delta y_i} = \sum_j p_{ij}\,L_{ij} - \sum_j q_{ij}\,L_{ij} = \mathcal{F}_{attr, i} - \mathcal{F}_{rep, i}
\label{eq:forces}
\end{equation}

\noindent $L_{ij}$ depends only on the pair-wise distances in the embedding, and the computation of the gradient represents a balance between attractive and repulsive forces exerted over each data point by all the others. On the one hand, high affinities in the HD space embedded as long distances in the LD space result in strong attractive forces, and low affinities in the HD space embedded as close distances result in weak attractive forces. On the other hand, repulsive forces monotonically decrease as a function of the embedding distance. Thus, the final embedding positions correspond to the point of equilibrium where attractive and repulsive forces are optimally balanced.

\paragraph{} In pt-SNE, the gradient descent is performed independently at each partial t-SNE, and the global cost function is computed as an average of the cost functions in each partial t-SNE,

\begin{equation*}
C = \frac{1}{z}\sum_{k} C^{k} = \frac{1}{z}\sum_{k} KL\left(P^{k}\|Q^{k}\right)
\end{equation*}

\paragraph{Finite affinity mass}
t-SNE holds an implicit dependence on the size $n$ of the data set. The reason is that the joint affinity distribution has a finite amount of probability mass to be allocated among all pairwise distances, the latter growing with $n\,\left(n-1\right)$. Therefore, as $n$ grows, affinities will be lower on average, tending to uniformity for the limiting case $n\rightarrow\infty$. This loss in discriminating power generates undesired side effects that we generically call the \textit{finite affinity mass} problem.

\paragraph{Pseudo-normalized cost function}

A first effect of the \textit{finite affinity mass} is that the cost function (Eq.~\ref{eq:cost_function}) holds itself an implicit dependence on $n$, and makes it difficult to objectively interpret its value. Taking into account that $p_{ij}$ and $q_{ij}$ decrease on average with $n\,\left(n-1\right)$, we see that this dependence amounts to,

\begin{flalign}
\nonumber
\langle C \rangle
&\propto -\log \langle q_{ij} \rangle \sum_{i,j} p_{ij} \\
&\propto \;\log\left(n \,\left(n-1\right)\right)
\label{eq:average_cost}
\end{flalign}

\noindent where we have dropped the term $\sum_{i,j}p_{ij}\log p_{ij}$, which is constant along the gradient descent.

The expression in Eq.~\ref{eq:average_cost} is the cost of a uniform distribution of affinities, i.e. the cost of a uniform embedding of $n\,\left(n-1\right)$ pairwise distances, expressing that all data points are equally similar. While it is not feasible to arrange $n\,\left(n-1\right)$ uniform pairwise distances in 2D, such a uniform distribution constitutes the worst possible embedding with respect to $P$, be $P$ what it may. Thus, Eq.~\ref{eq:average_cost} is an upper bound for $KL\left(P\|Q\right)$ and it makes sense to define a pseudo-normalized cost function as,

\begin{equation}
C = -\frac{\sum_{i,j}p_{ij}\log q_{ij}}{\log\left(n\,\left(n-1\right)\right)}
 = -\frac{H\left(P, Q\right)}{H\left(P, U\right)}
\label{eq:normalized_cost}
\end{equation}

\noindent In terms of information theory, this is the \textit{normalized cross-entropy} of distributions $P$ and $Q$, that is, the average cost of coding $P$ as $Q$ relative to the worst-case cost, which is the cost of a uniform embedding of $n\,\left(n-1\right)$ pairwise distances. Also, we can interpret this normalization factor as an expression of the loss in discriminating power due to the \textit{finite affinity mass} which, intuitively, we could approach as the relative increase of entropy of the affinity distribution, amounting to $\log\left(n\left(n -1\right)\right)$. This pseudo-normalized expression results in a value close to 1 for a random initial mapping, thus helping to make sense of the gradient descent and assess the stability of the final output.

\subsection*{Parametric configuration of the cost gradient}
\label{sec:parametric_configuration}

pt-SNE starts with a random embedding in a 2D half-unit disk (i.e. of radius r = 0.5) drawn from an isotropic Gaussian and updates embedding positions $y_{i}$ using the following expression,

\begin{flalign}
\nonumber
y^t_{i} 
&= y^{t-1}_{i} +\Delta y_{i}^t
\\
\Delta y_{i}^{d,t} 
&= \mu^t\Delta y_{i}^{d,t-1} - 4\,\eta^t\sum_{j\in nN_{i}}\left(\alpha^t\,p_{ij}-q^t_{ij}\right)\,\frac{l^{d,t}_{ij}}{1+\left(L^2_{ij}\right)^t}
\label{eq:mapping_update}
\end{flalign}

\noindent where $t, d$ are indicators of current iteration and embedding dimension respectively, $\mu$ is the \textit{momentum}, $\eta$ is the \textit{learning rate} and $\alpha$ is the \textit{exaggeration fator}. Also, we recall that $p_{ij}$ and $q_{ij}$ are the affinities in the HD/LD spaces, $L_{ij}$ is the pairwise distance in the embedding space and $l^d_{ij}$ is the component of $L{ij}$ along $d$.

\begin{figure}[!t]\centering
\includegraphics[width=15.0cm, height=5.3cm]{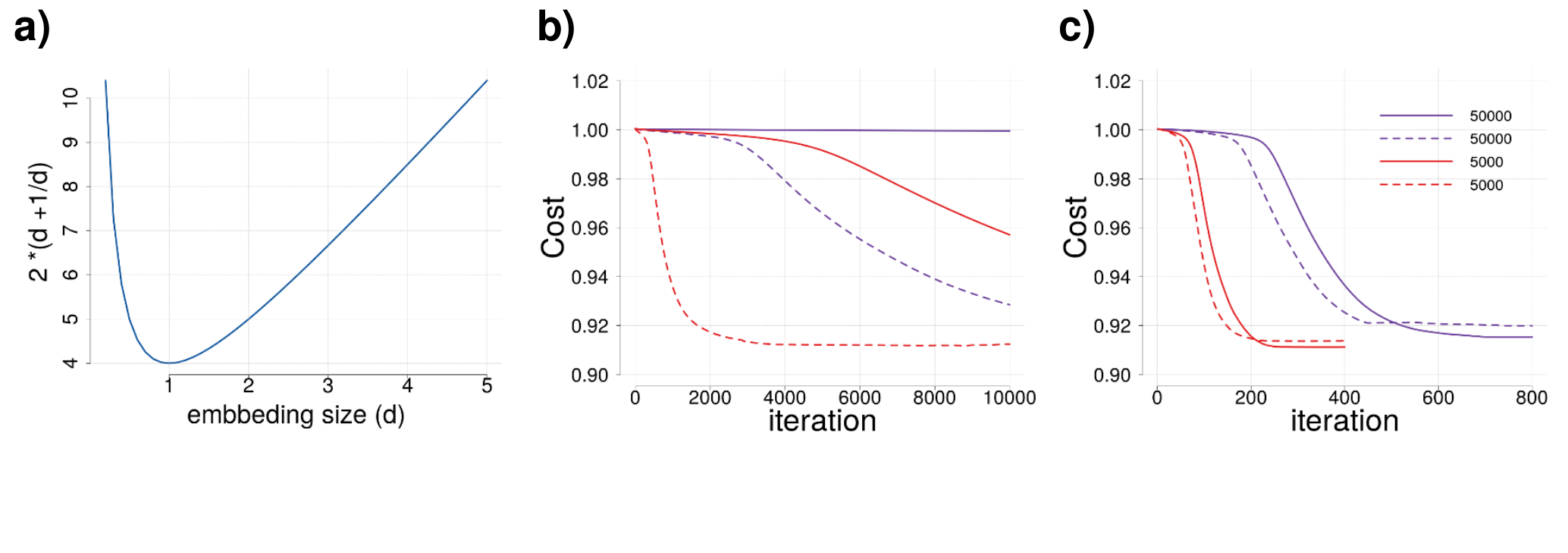}
\caption{\textbf{Effect of the parameters in the gradient descent}. a) Plain learning rate as resulting from Eq.\ref{eq:plain_learning_rate}; the learning-rate increases proportionally at both sides of $d=1$. b) Plain gradient descent as resulting from Eq.~\ref{eq:plain_learning_rate} (solid lines) and after correcting for sample size by a factor $\log\left(\nu\,N_{i}\right)$ (dashed lines). c) Gradient descent including gain $g\left(i, t\right)$ (Eq.~\ref{eq:learning_rate} (solid lines), and gradient descent with learning-rate (Eq.~\ref{eq:learning_rate}) and momentum (Eq.~\ref{eq:momentum}),  (dashed lines). Color indicates sample size.}
\label{fig:gradient_descent}
\end{figure}

\paragraph{Learning-rate} The learning-rate $\eta$ has a strong impact on convergence speed and it is not clear from the literature how to set its value: the default in most t-SNE implementations is $\eta=200$, while \cite{Wolf:2018} recommends increasing it to 1000 or \cite{Belkina:2019} suggests the value $\eta=n/12$. In pt-SNE, we use an auto-adaptive scheme given by,

\begin{equation}
\eta^t_{i} = 2\,\left(d^t+\frac{1}{d^t}\right)\,\log\left(\nu\,N^k_{i}\right)\,g^t_{i}
\label{eq:learning_rate}
\end{equation}

where $d^t$ is the diameter of the embedding at iteration $t$, and where we note the following:

\begin{itemize}

\item
While the size of the embedding clearly changes along the optimization, Eq.~\ref{eq:mapping_update} seems to be lacking a reference size for the distance factors $L_{ij}$ and $l_{ij}$, thus suggesting,

\begin{equation*}
\eta^t \propto \frac{1+\left(d^t\right)^2}{d^t} = d^t+\frac{1}{d^t} 
\end{equation*}

The factor $\left(d^t+\frac{1}{d^t}\right)$ renders Eq.\ref{eq:mapping_update} independent of the size of the embedding and, interestingly, yields equally increasing learning-rates at both sides of $d=1$ (Fig.~\ref{fig:gradient_descent}~a) i.e. either expanding or compressing embedding, the latter occurring for large ppx. Furthermore, if we specifically set

\begin{equation}
\eta^t \propto 2\,\frac{1+\left(d^t\right)^2}{d^t} = \frac{1+\left(d^t\right)^2}{r^t}
\label{eq:plain_learning_rate}
\end{equation}

\noindent where $r$ is the radius of the embedding, then for $d=1$ we have $\eta=4$, so that we make sense out of factor 4 in the gradient expression (Eq. \ref{eq:cost_gradient}) as the learning-rate corresponding to a half-unit disk embedding (i.e. $r=0.5$). Therefore we can drop the factor 4 from Eq.~\ref{eq:mapping_update}.

\item
$\log\left(\nu\,N^k_{i}\right)$ compensates the effect of the \textit{finite affinity mass} problem rendering Eq.\ref{eq:mapping_update} independent of the size of the affinity matrix operating on each partial t-SNE, actually given by $\nu\,N^k_i =\left(\rho\,n\right)\,\left(\rho\,3\,ppx\right)$.

\item
$g^t_i$ is an acceleration factor given as a point-wise step size update by means of the Jacobs adaptive learning rate scheme \cite{Jacobs:1988} which is indeed implemented in BHt-SNE \cite{Maaten:2014}. This factor increases the learning rate in directions in which the gradient is stable, anticipating position updates that are likely to occur in the next steps,

\begin{flalign*}
\text{init:}\quad &g\left(i, 1\right) = 1.0
\\
\text{if}\quad & \text{gradient\_direction}\left(i, t\right) == \text{gradient\_direction}\left(i, t-1\right)
\\
&g\left(i, t\right) \,+= 0.1\,\text{gain}
\\
\text{else}\quad &
\\
&g\left(i, t\right) \,*= \left(1 - 0.1\,\text{gain}\right)
\end{flalign*}

\noindent with $gain = 2.0$ by default.
\end{itemize}

In summary, our auto-adaptive scheme for the learning-rate (Eq.\ref{eq:learning_rate}) operates in favor of a stable gradient descent. On the initial stages of the optimization, the mismatch between $p_{ij}$ and $q_{ij}$ will likely be large but the embedding size is small, thus the learning-rate will mitigate the impact of the strong attraction/repulsion forces originated during this critical moment. Along the course of the optimization, the size of the embedding will get larger, increasing the learning rate to compensate the decreasing attraction/repulsion forces.

\paragraph{Momentum} Using momentum $\mu$ speeds up the gradient descent but might blow up the embedding size on the final steps. In this final stage, the learning rate is high because the embedding has grown large, but the attractive/repulsive forces are almost balanced hence the position updates should be minor. Therefore adding momentum is risky. Because of this, we use a decaying momentum of the form,

\begin{equation}
\mu^t = 0.8\,\left(1-\frac{e}{\epsilon}\right)^2
\label{eq:momentum}
\end{equation}

\noindent where $e$ stands for the current epoch and $\epsilon$ is the total number of epochs.

\paragraph{Exaggeration factor} Using the exaggeration factor $\alpha$ is likely to be detrimental in pt-SNE. Forcing an early arrangement of clusters in each thread might depict a more fragmented global solution at the early stages of the gradient descent, which is unlikely to be solved in subsequent epochs. Therefore we dismiss using exaggeration in pt-SNE.

\paragraph{} We illustrate the effect of the parametric setup described above in the gradient descent (Fig.\ref{fig:gradient_descent}~b,~c). We note that a plain learning-rate (Eq.\ref{eq:plain_learning_rate}) yields a very smooth decay of the embedding cost (Fig.~\ref{fig:gradient_descent}~b, solid lines) with a strong dependence on the size of the data set $n$. Including the acceleration factor $g\left(i, t\right)$ has a massive impact on the gradient descent, dramatically reducing the number of iterations needed to reach a stable embedding (Fig.~\ref{fig:gradient_descent}~c, compare the range of the x-axis with panel b). Using a decaying momentum contributes to faster reaching a stable solution, although in general, the final embedding is not so good (Fig.\ref{fig:gradient_descent}~c dashed lines).

\section*{Data availability}

All data used in this paper is available at https://github.com/jgarriga65/bigMap/examples.

\section*{Code availability}

The pt-SNE algorithm is implemented in the \textbf{bigMap} R-package along with additional tools for the post-processing of the output. We worked out the examples in this paper using version bigMap\_4.5.6 available at https://github.com/jgarriga65/bigMap/package and using the HPC platform at the Computational Biology Lab (CEAB-CSIC) (Table~\ref{tbl:CBLab}). 

\section*{Acknowledgments}
We acknowledge members from the Theoretical and Computational Ecology lab to provide comments on previous versions of the MS, and insights as beta users of the bigMap R-package. This work was supported by the Spanish Ministry (MINECO, Grant CGL2016-78156-R), and the Max Planck Institute for Ornithology (MPIO, Germany). The high-performance computation cluster at the Computational Biology Lab (CEAB-CSIC) was supported by the Spanish Ministry (MINECO, CSIC13-4E-1999).

\section*{Author information}
The authors contributed equally in designing and writing the paper. J.G developed pt-SNE and performed the analysis.

\section*{Competing interests}
The authors declare no competing interests.

\begin{table}[!t]
\centering
\small
\begin{tabular}{ccccccc}
nodes & model & CPU & cores & fr.(MHz) & RAM & OS (bits)\\
\hline
5 & PowerEdge R420 & Intel(R) Xeon(R) E5-2450L & 16 & 1800 & 161G & 64 \\
7 & PowerEdge R430 & Intel(R) Xeon(R) E5-2650 & 20 & 2300 & 193G & 64 \\
1 & PowerEdge R815 & AMD Opteron(tm) 6380 & 64 & 2500 & 515G & 64 \\
\hline
\end{tabular}
\caption{\textbf{High-performance computing cluster at the Computational Biology Lab (CEAB-CSIC)}. Technical specifications.}
\label{tbl:CBLab}
\end{table}

\bibliography{main.bib}

\newpage

\section*{Supplementary information}

\section{Retrieval-information interpretation of t-SNE}
\label{sec:retrieval_information}

Assessing the quality of a low-dimensional representation of a data set independently from a method's inherent criteria is not a trivial task, and has inspired the development of different measures \textit{e.g.} \cite{Lueks:2011, Lee:2009, Kaski:2003}. Among them, \cite{Venna:2010} give a rigorous definition of the quality of an embedding from an information retrieval perspective.

Recalling t-SNE, the affinities between data points are represented as joint distributions $P$ and $Q$ in both the high and low dimensional spaces,

\begin{flalign*}
p_{j\mid i} &= \frac{\exp{\left(-\beta_{i}\,\|x_i-x_j\| ^2\right)}}{\sum_{k\neq i}\exp{\left(-\beta_{i}\,\|x_i-x_k\|^2\right)}}
\\
p_{ij} &= \frac{p_{j\mid i}+p_{i\mid j}}{2n} = p_{ji}
\\
q_{ij} &= \frac{\left(1 + \| y_i-y_j \|^2\right)^{-1}}{\sum_{k\neq l}\left(1 + \| y_k-y_l \|^2\right)^{-1}}
\end{flalign*}

\noindent and a gradient descent is performed to find a low-dimensional representation of the data that minimizes the mismatch between $p_{ij}$ and $q_{ij}$, with a cost function based on the Kullback-Leibler divergence,

\begin{equation*}
C = KL\left(P\|Q\right)=\sum_{i,j}p_{ij}\log\frac{p_{ij}}{q_{ij}}
\end{equation*}

Following \cite{Venna:2010}, the conditional similarities in the high dimensional space $p_{j|i}$ define a \textit{probabilistic model of prevalence} for the neighbors of $\mathbf{x}_{i}$ while the conditional similarities in the low dimensional space $q_{j|i}$ define a \textit{probabilistic model of retrieval} for the neighbors of $\mathbf{y_{i}}$. Hence, natural measures of matching between both models arise from the divergences,

\begin{flalign*}
D\left(p_{.\mid i}, q_{.\mid i}\right) &= \sum_{j\neq i}p_{j|i}\log\frac{p_{j|i}}{q_{j|i}} \\
D\left(q_{.\mid i}, p_{.\mid i}\right) &= \sum_{j\neq i}q_{j|i}\log\frac{q_{j|i}}{p_{j|i}}
\end{flalign*}

\noindent These two measures are called \textit{smoothed recall} and \textit{smoothed precision} \cite{Venna:2010}, respectively, as they are intrinsically related to and represent a generalization of \textit{recall} and \textit{precision} for probabilistic neighborhoods: the fundamental trade-off between precision and recall in information retrieval is expressed here as a trade-off between missing similar data points versus retrieving dissimilar data points.

Pushing this argument further, the t-SNE cost function yields the following interpretation,

\begin{flalign}
\nonumber
KL\left(P\|Q\right)
&= \sum_{i,j}p_{ij}\log\frac{p_{ij}}{q_{ij}}
\\
\nonumber
&= \sum_{i, j}p_{j\mid i}\,p_{i}\,\log\frac{p_{j\mid i}\,p_{i}}{q_{j\mid i}\,q_{i}}
\\
\nonumber
&= \sum_{i, j}p_{j\mid i}\,p_{i}\,\left(\log\frac{p_{j\mid i}}{q_{j\mid i}} +\log\frac{p_{i}}{q_{i}}\right)
\\
\label{eq:wRecall}
&= \sum_{i}p_{i}\sum_{j\neq i}p_{j\mid i}\log\frac{p_{j\mid i}}{q_{j\mid i}} +\sum_{i}p_{i}\log\frac{p_{i}}{q_{i}}
\end{flalign}

\noindent with,

\begin{flalign}
\label{eq:ld_global_prevalence}
q_{i} &= \sum_{j\neq i}q_{ij}=\sum_{i\neq j}q_{ji}\,,\quad\sum_{i}q_{i}=1
\\
\label{eq:hd_global_prevalence}
p_{i} &= \sum_{j\neq i} p_{ij} = \frac{1}{2\,n}+ \frac{1}{2\,n}\,\sum_{k\neq i}p_{i|k}\,,\quad\sum_{i}p_{i}=1
\end{flalign}

\noindent and,

\begin{displaymath}
p_{i|k} = \sum_{k\neq i} \frac{\exp{\left(-\beta_{k}\,d_{ki}^2\right)}}{\sum_{l\neq k}\exp{\left(-\beta_{k}\,d_{kl}^2\right)}}
\end{displaymath}

\noindent The distributions $p_{i}$ and $q_{i}$ (Eqs. \ref{eq:ld_global_prevalence} and \ref{eq:hd_global_prevalence}) express the overall probability of picking $\textbf{x}_{i}$ as a neighbor of any of the rest of the data points and therefore they can be regarded as \textit{probabilistic models of global prevalence} in the high and low dimensional spaces respectively. Thus,

\begin{itemize}
\item the first term in Eq.~\ref{eq:wRecall}, $\mathbb{E}_{p_{i}}\left[D\left(p_{.\mid i}, q_{.\mid i}\right)\right]$, is the \textit{expected smoothed recall} with respect to $p_{i}$, a weighted average \textit{smoothed recall}, where the weights $w_{i}\equiv p_{i}$ are such that $w_{i}>w_{k}$ if the neighborhood of $\textbf{x}_{i}$ is denser than the neighborhood of $\textbf{x}_{k}$;
\item the second term in Eq.~\ref{eq:wRecall}, $D\left(p_{i}, q_{i}\right)$, is a measure of matching between both models of global prevalence that can be interpreted as a measure of \textit{global structure preservation}.
\end{itemize}

In summary, Eq.~\ref{eq:wRecall} yields an interesting insight about the cost function from a retrieval information perspective: the t-SNE approach maximizes the \textit{expected smoothed recall}, giving priority to areas where local structure is more significant while trying to preserve the global prevalence of the data points in the high dimensional space, what we can rewrite as,

\begin{equation*}
K\left(P\| Q\right) \equiv \mathbb{E}_{p_{i}}\left[D\left(p_{.\mid i}, q_{.\mid i}\right)\right] + D\left(p_{i\mid .}, q_{i\mid .}\right)
\end{equation*}

\noindent where we have changed notation ($p_{i\mid .}\equiv p_{i}, q_{i\mid .}\equiv q_{i}$) to stress the relation of each term with the preservation of local and global structure respectively.

\section{Affinities}

Given a random variable $X$ and a model $P$ of the probability distribution of $X$ with entropy $H\left(P\right)$, the perplexity of $P$ is a measure of how well the model predicts the outcome of $X$, defined as:

\begin{equation}
 Perplexity\left(P\right)=2^{H\left(P\right)}
\label{eq:perplexity_supp}
\end{equation}

In the context of the t-SNE, $P$ is called affinity distribution and represents the probability of picking neighbouring datapoints based on pair-wise similarity. Although different measures of similarity can be considered, the most commonly used is the Euclidean distance. Given a data set $X=\{\mathbf{x}_{1},\dots,\mathbf{x}_{n}\}$, the t-SNE transforms \textit{similarities} to $x_{i}$ (whatever measure of similarity) into a probability distribution of picking each one of the data points as a neighbor of $x_{i}$. This probability distribution is determined by placing a \textit{kernel} centered at $x_{i}$,

\begin{equation*}
 \mathcal{K}_{i}\equiv\mathcal{K}\left(\frac{\mathbf{x}_{i}-\mathbf{x}_{j}}{h_{i}}\right),\quad\forall\;x_{j}\in X
\end{equation*}

\noindent where $\mathcal{K}\left(\cdot\right)$ is a non-negative function that integrates to one and has zero mean, and $h_{i}>0$ is a smoothing parameter called the \textit{bandwidth} \cite{Terrell:1992}. A kernel function can be \textit{fixed}, in which case $h_{i}=h$ for any $\mathbf{x}_{i}\in X$, or \textit{adaptive} where the bandwidth is dependent on $x_{i}$. Using one or the other has particular benefits and drawbacks depending on the final objective \cite{Terrell:1992}.

In particular, t-SNE transforms distances $d_{ij}=\|\mathbf{x}_{i}-\mathbf{x}_{j}\|$ by means of an adaptive Gaussian kernel defined as,

\begin{equation*}
 \mathcal{K}_{ij}=\frac{1}{C_{i}}\,\exp\left(-\beta_{i}\,d_{ij}^2\right)
\end{equation*}

\noindent with bandwidth $h_{i}=1/\sqrt{\beta_{i}}$ and where $C_{i}=\sum_{j}\exp\left(-d_{ij}^2\,\beta_{i}\right)$ is a normalization constant.

\subsection{Computation of the kernel bandwidths}

We use Equation \ref{eq:perplexity_supp} to derive the bandwidth $1/\sqrt{\beta_{i}}$ such that yields a given value of perplexity $\upsilon$,

\begin{equation}
 \log_2\left(\upsilon\right)=-H\left(\mathcal{K}_{i}\right)=-\sum_{j}\mathcal{K}_{ij}\,\log_2\left(\kappa_{ij}\right)
\label{eq:bandwidth}
\end{equation}

Equation \ref{eq:bandwidth} does not have a close form solution but the value of $\beta_{i}$ can be approximated using a binary search upon a range of values for $\beta_{i}$. To efficiently compute Equation \ref{eq:bandwidth} we use,

\begin{flalign*}
 -\sum_{j}\mathcal{K}_{ij}\,\log_2\left(\mathcal{K}_{ij}\right)
 &=-\sum_{j}\frac{1}{C}\,\exp\left(-d_{ij}^2\,\beta_{i}\right)\,\log_2\left(\frac{1}{C}\,\exp\left(-d_{ij}^2\,\beta_{i}\right)\right)
 \\
 &=\frac{-1}{C}\,\sum_{j}\exp\left(-d_{ij}^2\,\beta_{i}\right)\,\left(\frac{-d_{ij}^2\,\beta_{i}}{\ln\left(2\right)}-\log_2\left(C\right)\right)
 \\
 &=\frac{-1}{C}\left(\sum_{j}\frac{-d_{ij}^2\,\beta_{i}}{\ln\left(2\right)}\,\exp\left(-d_{ij}^2\,\beta_{i}\right)-\log_2\left(C\right)\,\sum_{j}\exp\left(-d_{ij}^2\,\beta_{i}\right)\right)
 \\
 &=\log_2\left(C\right)+\frac{\beta_{i}}{C\,\ln\left(2\right)}\,\sum_{j}d_{ij}^2\,\exp\left(-d_{ij}^2\,\beta_{i}\right)
\end{flalign*}

Thus,

\begin{equation}
 \ln\left(\upsilon\right)=\ln\left(C\right)+\frac{\beta_{i}}{C}\,\sum_{j}d_{ij}^2\,\exp\left(-d_{ij}^2\,\beta_{i}\right)
\label{eq:bandwidthFast}
\end{equation}

\subsection{A reformulation of perplexity}

Let's denote $\nu \equiv Perplexity\left(P_{.\mid i}\right)$ defined as,

\begin{equation*}
\nu = 2^{H\left(P_{.\mid i}\right)}
\end{equation*}

\noindent where,

\begin{equation*}
H\left(P_{.\mid i}\right) = -\,\sum_{j \in N_i}\,p_{j\mid i}\,log\,p_{j\mid i}
\end{equation*}

\noindent thus,

\begin{equation}
log\,\nu = -\,\sum_{j \in N_i}\,p_{j\mid i}\,log\,p_{j\mid i}
\label{eq:logppx}
\end{equation}

Given the input affinity distribution,

\begin{equation*}
p_{j\mid i} = \frac{\exp{\left(-\beta_{i}\,\|x_i-x_j\| ^2\right)}}{ \sum_{k\neq i}\exp{\left(-\beta_{i}\,\|x_i-x_k\|^2\right)}}
\end{equation*}

\noindent denote,

\begin{flalign*}
exp_{ji} &\equiv \exp{\left(-\beta_{i}\,\|x_i-x_j\| ^2\right)}\,,
\\
C_{i} &\equiv \sum_{k\neq i}\exp{\left(-\beta_{i}\,\|x_i-x_k\|^2\right)}
\end{flalign*}

\noindent Then, plugging the input affinity distribution into Eq.\ref{eq:logppx} we have,

\begin{flalign*}
log\,\nu 
&= -\,\sum_{j \in N_i}\,\frac{exp_{ji}}{Ci}\,log\,\frac{exp_{ji}}{Ci}
\\
&= -\,\sum_{j \in N_i}\,\frac{exp_{ji}}{Ci}\,\left(log\,exp_{ji}-log\,Ci\right)
\\
&= -\,\sum_{j \in N_i}\,\frac{exp_{ji}}{Ci}\,log\,exp_{ji}+\sum_{j \in N_i}\,\frac{exp_{ji}}{Ci}\,log\,Ci
\\
&= -\,\sum_{j \in N_i}\,p_{j\mid i}\,\left(-\beta_{i}\,\|x_i-x_j\| ^2\right)+log\,Ci\sum_{j \in N_i}\,p_{j\mid i}
\\
&= \beta_{i}\,\sum_{j \in N_i}\,p_{j\mid i}\,\|x_i-x_j\| ^2+log\,Ci
\\
&= \beta_{i}\,E_{P.\mid i}\left[\,\|x_i-x_j\|^2\right]+log\,Ci
\end{flalign*}

Thus,

\begin{flalign}
\nonumber
log\,Ci-log\,\nu 
&= -\beta_{i}\,E_{P.\mid i}\left[\,\|x_i-x_j\|^2\right]
\\
\nonumber
log\,\frac{C_{i}}{\nu}
&= -\beta_{i}\,E_{P.\mid i}\left[\,\|x_i-x_j\|^2\right]
\\
\nonumber
\frac{C_{i}}{\nu}
&= \exp{\left(-\beta_{i}\,E_{P.\mid i}\left[\,\|x_i-x_j\|^2\right]\right)}
\\
\label{eq:ppx_reformulated}
\frac{1}{\nu}
&= \frac{1}{C_{i}}\,\exp{\left(-\beta_{i}\,E_{P.\mid i}\left[\,\|x_i-x_j\|^2\right]\right)}
\end{flalign}

That is, perplexity is the inverse of the density of the expected squared distance given the kernel itself or, in other words, the higher the perplexity the lower the density associated to the expected squared distance.

\subsection{Special case: uniform affinity distribution}
\label{sec:lambertW}

A special case that may arise (in particular when measuring similarities in discrete data sets using Manhattan distances or alike) is to have all data points in the neighborhood of $x_{i}$ at the same distance. This is the case, for instance, in the Primes1M data set~\cite{Williamson:2019}, which we use as an example in this work. In such situations, we have that,

\begin{itemize}
\item $\forall j\in N_{i},\, \|x_i-x_j\|^2 \equiv E_{i}$;
\item the affinity distribution $P_{.\mid i}$ is uniform and independent of $E_{i}$;
\item the perplexity is exactly the number of neighbors,
\begin{flalign*}
E_{P.\mid i}\left[\,\|x_i-x_j\|^2\right] 
& = E_{i}
\\
\mbox{thus, }\quad \frac{1}{\nu} 
&= \frac{\exp{\left(-\beta_{i}\,E_{i}\right)}}{|N_{i}|\,\exp{\left(-\beta_{i}\,E_{i}\right)}}\,;
\end{flalign*}
\item more importantly, the bandwidth $\beta_{i}$ is undetermined (Eq.\ref{eq:ppx_reformulated}.
\end{itemize}

In this case we assume a Gaussian kernel where the expected squared distance is exactly $E_{i}$ so that Eq.~\ref{eq:ppx_reformulated} becomes,

\begin{flalign*}
\frac{1}{\nu} 
&= \sqrt{\frac{\beta_{i}}{\pi}}\,\exp{\left(-\beta_{i}E_{i}\right)}
\\
\mbox{with}\quad \beta_{i} 
&= \frac{1}{2\sigma^2_{i}}
\end{flalign*}

\noindent which can otherwise be written as,

\begin{equation*}
\exp{\left(2\beta_{i}E_{i}\right)} = \beta_{i}\,\frac{\nu^2}{\pi}
\end{equation*}

\noindent and making $x\equiv 2\beta_{i}E_{i}$,

\begin{flalign*}
\exp{\left(x\right)} 
&= x\frac{\nu^2}{2\pi E_{i}}
\\
-x\,\exp{\left(-x\right)}
&= -\frac{2\pi E_{i}}{\nu^2}
\end{flalign*}

\noindent gives a solution for $\beta_{i}$ by means of the Lambert $\mathcal{W}$ function,

\begin{flalign*}
-x &= \mathcal{W}_{1}\left(-\frac{2\pi E_{i}}{\nu^2}\right)
\\
\mbox{thus,}\quad -2\beta_{i}E_{i}
&= \mathcal{W}_{1}\left(-\frac{2\pi E_{i}}{\nu^2}\right)
\\
\beta_{i} 
&= \frac{-1}{2\,E_{i}}\,\mathcal{W}_{1}\left(-\frac{2\pi E_{i}}{\nu^2}\right)
\end{flalign*}

\section{Accurately exploring global and local data structure}
\label{sec:combo}

Classic implementations of t-SNE accelerate the computation of attractive forces by limiting the size of the nearest neighbors' sets ($N_{i}$) to $|N_{i}|=3\,ppx$ \cite{Maaten:2009a}. In our case, this same approach, combined with low thread-ratios, can produce blurred visualizations (compare Fig.~4~a with Fig.~\ref{fig:10xm_comp}~a). The reason for this is that the final mapping positions result from the partial neighborhoods considered in the last epoch, involving just a fraction $\rho$ of the whole neighborhood  ($|N^k_{i}|\approx\rho\,N_{i}$, see section Methods). In most cases, this is not an issue, and pt-SNE will do as a stand-alone methodology. Nonetheless, we can get improved visualizations by combining pt-SNE with a subsequent algorithm, using the output of the former to initialize the latter.  Typically, we would run pt-SNE at large ppx to capture the global structure, followed by FIt-SNE at small ppx to refine the local one. Starting with an informative initialization, a few iterations without data chunking and at a small ppx value suffice to refine local data structures and get more appealing visualizations. Thus, this post-processing should not take a long time.

\begin{figure}[tp!]
\centering
\includegraphics[width=13.0cm, height=8.0cm]{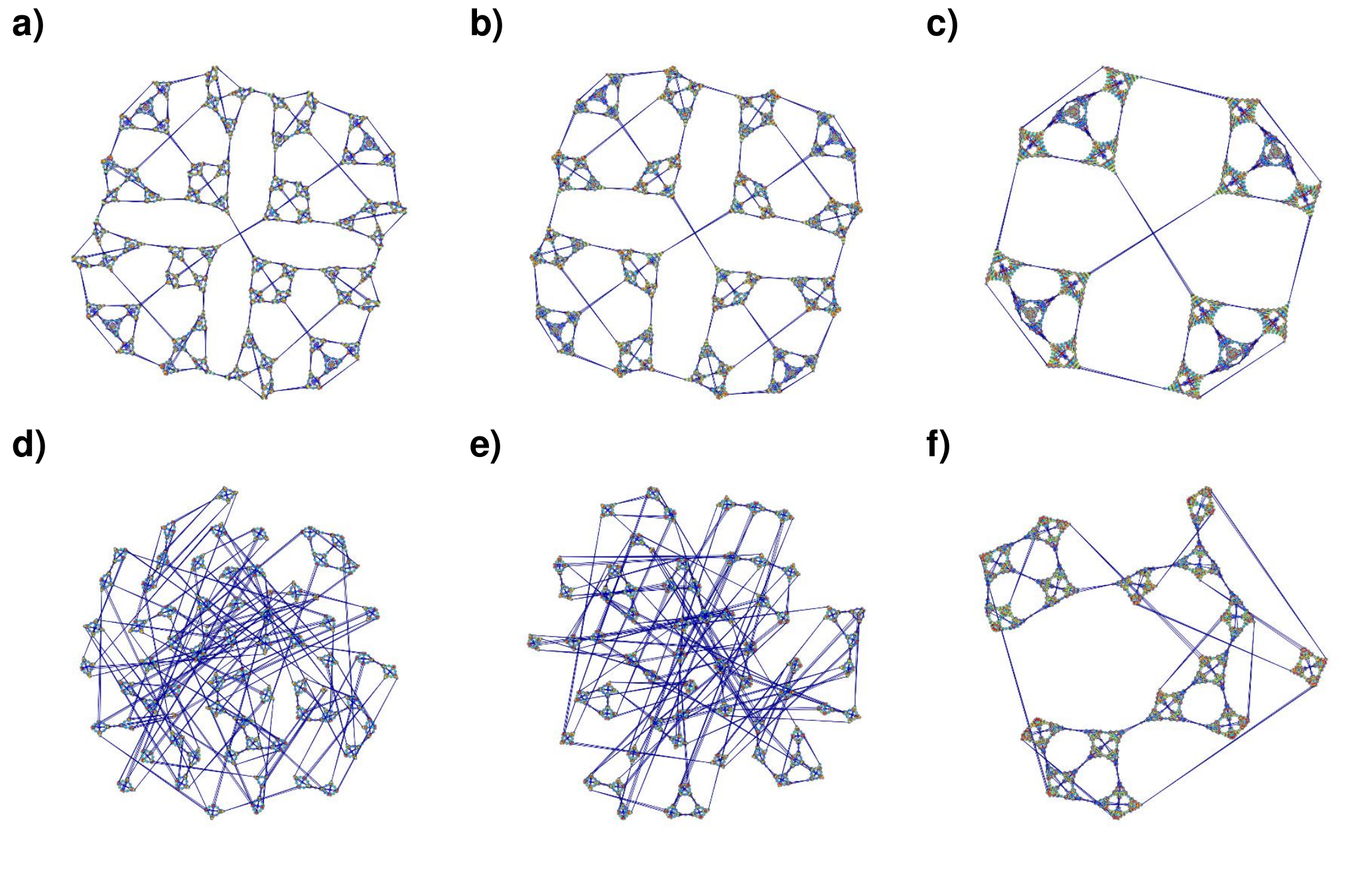}
\caption{\textbf{Combining pt-SNE and FIt-SNE} We used the pt-SNE embedding shown in Fig.~2~f ($ppx = 2030$, thread-ratio $\rho=0.25$) to initialize several runs of FIt-SNE with: a) $ppx=10$, b) $ppx =20$, c) $ppx=102$. d), e), and f): For comparative purposes we reproduce the outputs of pt-SNE with the values of ppx used in a, b and c, but using random initialization (i.e. the same procedure shown in Fig.~1, panels a, b, and c). Note that while both rows are showing the same scales of structure, in the upper row all pieces of the structure are correctly placed, reproducing also the global structure injected from the start. Colors depict relative pair-wise distances in original space (red:closer, blue:farther)}
\label{fig:combo1}
\end{figure}

\begin{figure}[tp!]
\centering
\includegraphics[width=14.3cm, height=6.25cm]{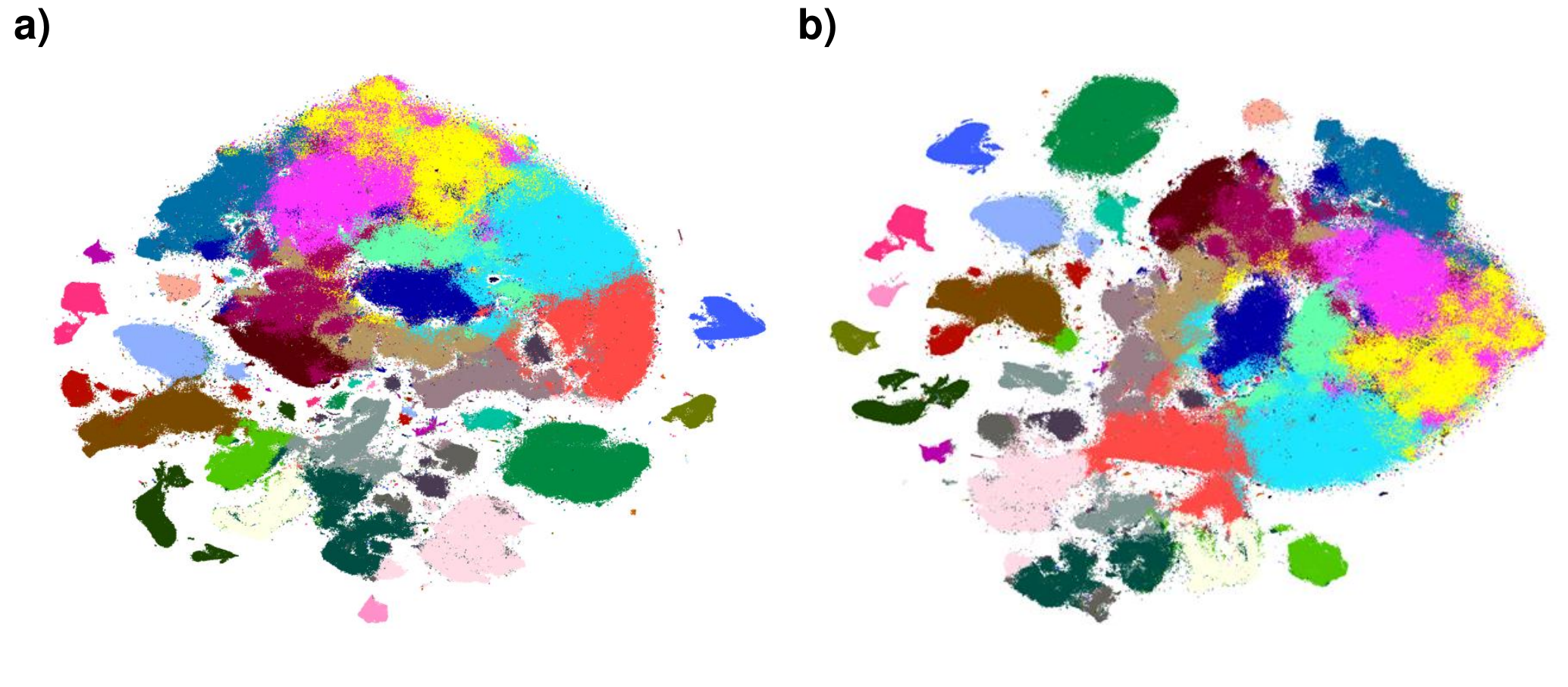}
\caption{\textbf{Using pt-SNE to initialize FIt-SNE}. 1.3 Million Brain Cells from E18 Mice \cite{10xGenomics:2018}. Clustering labels taken from \cite{Wolf:2018}. a) Combining pt-SNE with $ppx=65306$ and FIt-SNE with $ppx=130$, 5.37 h running time; b) FIt-SNE embedding with $ppx=260$ and PCA initialization, 1.06 h running time.}
\label{fig:10xm_comp}
\end{figure}

The Sierpinski-3D data set perfectly illustrates the effect of this combination (Fig.~\ref{fig:combo1}). The embedding in Fig.~2~f (i.e. using high perplexity and low thread-ratio) showed a good visualization of global structure but a high loss of local accuracy. We used this embedding to start several runs of FIt-SNE with different perplexities. The result (Fig.~\ref{fig:combo1}~a,~b,~c) is that the local structure at the corresponding scale is enhanced while the global structure is mostly maintained. For comparative purposes we show the corresponding visualizations (Fig.~\ref{fig:combo1}~d, ~e, ~f) that resulted from the procedure used in Fig.1 (i.e. using random initialization).

We show a further example with the \cite{10xGenomics:2018} data set where we combined pt-SNE ($ppx=65306$) and FIt-SNE ($ppx=130$) (Fig.\ref{fig:10xm_comp}~a), using the output of the former (Fig.~4~a) to initialize the later, and we compare it with a single run of FIt-SNE with $ppx=260$ and standard configuration (PCA initialization, learning-rate $\eta=n/12$, early-exaggeration $\alpha=12$, Fig.\ref{fig:10xm_comp}~b). The similarity between both results is clear. However, the kNP values were $linAUC=.5029,\,logAUC=.2047$ for the pt-SNE+FIt-SNE combined strategy, and $linAUC=.4742,\,logAUC=.2515$ for the FIt-SNE alone. Thus, we improve the accuracy at global scales while minimizing the loss of local structure.

\end{document}